\documentclass[letterpaper, 10 pt, journal, twoside]{IEEEtran}  

\usepackage{etex}

\usepackage{multicol}
\usepackage{hyperref}

\usepackage{amsmath,amsfonts,amssymb,amsthm} 
\usepackage{float}
\usepackage{bm}
\usepackage{graphicx}
\usepackage{epstopdf}
\usepackage{epsfig}
\usepackage{xspace}
\usepackage{multirow}
\usepackage{hhline}
\usepackage[export]{adjustbox}
\usepackage{csvsimple}

\usepackage{graphicx}
\usepackage[colorinlistoftodos]{todonotes}
\newcommand{\journalVersion}[1]{}
\usepackage{xr}

\usepackage{comment}
\usepackage{siunitx}
\usepackage{relsize}
\usepackage{ifthen}

\usepackage[caption=false]{subfig}

\usepackage[vlined,ruled,linesnumbered]{algorithm2e}
\usepackage{graphics} 
\usepackage{rotating}
\usepackage{color}
\usepackage{enumerate}
\usepackage[T1]{fontenc}
\usepackage{psfrag}
\usepackage{epsfig} 
\usepackage{booktabs}
\usepackage{graphicx,url}
\usepackage{multirow}
\usepackage{array}
\usepackage{latexsym}
\usepackage{amsfonts}
\usepackage{amsmath}
\usepackage{amssymb}
\usepackage{xstring}
\usepackage[noend]{algorithmic}
\usepackage{multirow}
\usepackage{xcolor}
\usepackage{prettyref}
\usepackage{flexisym}
\usepackage{bigdelim}
\usepackage{breqn} 
\usepackage{listings}
\let\labelindent\relax
\usepackage{enumitem}
\usepackage{xspace}
\usepackage{bm}
\graphicspath{{./figures/}}
\usepackage{tikz}
\usetikzlibrary{matrix,calc}

\usepackage{mdwlist}

\let\itemize\undefined
\makecompactlist{itemize}{stditemize}

\newrefformat{prob}{Problem\,\ref{#1}}
\newrefformat{def}{Definition\,\ref{#1}}
\newrefformat{sec}{Section\,\ref{#1}}
\newrefformat{sub}{Section\,\ref{#1}}
\newrefformat{prop}{Proposition\,\ref{#1}}
\newrefformat{app}{Appendix\,\ref{#1}}
\newrefformat{alg}{Algorithm\,\ref{#1}}
\newrefformat{cor}{Corollary\,\ref{#1}}
\newrefformat{thm}{Theorem\,\ref{#1}}
\newrefformat{lem}{Lemma\,\ref{#1}}
\newrefformat{fig}{Fig.\,\ref{#1}}
\newrefformat{tab}{Table\,\ref{#1}}

\newtheorem{theorem}{Theorem}

\newtheorem{proposition}[theorem]{Proposition}

\newcommand{\cf}{\emph{cf.}\xspace}

\newcommand{\bdmath}{\begin{dmath}}
\newcommand{\edmath}{\end{dmath}}
\newcommand{\beq}{\begin{equation}}
\newcommand{\eeq}{\end{equation}}
\newcommand{\bdm}{\begin{displaymath}}
\newcommand{\edm}{\end{displaymath}}
\newcommand{\bea}{\begin{eqnarray}}
\newcommand{\eea}{\end{eqnarray}}
\newcommand{\beal}{\beq \begin{array}{ll}}
\newcommand{\eeal}{\end{array} \eeq}
\newcommand{\beas}{\begin{eqnarray*}}
\newcommand{\eeas}{\end{eqnarray*}}
\newcommand{\ba}{\begin{array}}
\newcommand{\ea}{\end{array}}
\newcommand{\bit}{\begin{itemize}}
\newcommand{\eit}{\end{itemize}}
\newcommand{\ben}{\begin{enumerate}}
\newcommand{\een}{\end{enumerate}}

\newcommand{\calB}{{\cal B}}
\newcommand{\calC}{{\cal C}}

\newcommand{\calE}{{\cal E}}

\newcommand{\calG}{{\cal G}}

\newcommand{\calP}{{\cal P}}

\newcommand{\calU}{{\cal U}}
\newcommand{\calV}{{\cal V}}

\newcommand{\etal}{\emph{et~al.}\xspace}
\newcommand{\setal}{~\emph{et~al.}\xspace}

\newcommand{\M}[1]{{\bm #1}} 
\renewcommand{\boldsymbol}[1]{{\bm #1}}

\newcommand{\hide}[1]{}

\newcommand{\hiddenText}{{\color{gray} hidden text.}}
\newcommand{\hideWithText}[1]{\hiddenText}

\newcommand{\subject}{\text{ subject to }}

\newcommand{\lone}{\ell_{1}}

\newcommand{\tran}{^{\mathsf{T}}}

\newcommand{\trace}[1]{\mathrm{tr}\left(#1\right)}

\newcommand{\rank}[1]{\mathrm{rank}\left(#1\right)}

\newcommand{\zero}{{\mathbf 0}}
\newcommand{\eye}{{\mathbf I}}

\newcommand{\matTwo}[1]{\left[\begin{array}{cc}  #1  \end{array}\right]}

\newcommand{\Real}[1]{ { {\mathbb R}^{#1} } }

\newcommand{\SO}[1]{\ensuremath{\mathrm{SO}(#1)}\xspace}

\newcommand{\MA}{\M{A}}

\newcommand{\MM}{\M{M}}
\newcommand{\MN}{\M{N}}

\newcommand{\MQ}{\M{Q}}
\newcommand{\MU}{\M{U}}
\newcommand{\MR}{\M{R}}

\newcommand{\MI}{\M{I}}

\newcommand{\MT}{\M{T}}
\newcommand{\MX}{\M{X}}

\newcommand{\MW}{\M{W}}
\newcommand{\MZ}{\M{Z}}
 
\newcommand{\MOmega}{\M{\Omega}}

\newcommand{\vt}{\boldsymbol{t}}

\renewcommand{\ij}{_{ij}}

\newcommand{\scenario}[1]{{\smaller \sf#1}\xspace}

\newcommand{\vertigo}{\scenario{Vertigo}}

\newcommand{\cvx}{{\sf cvx}\xspace}

\newcommand{\blue}[1]{{\color{blue}#1}}

\newcommand{\linkToPdf}[1]{\href{#1}{\blue{(pdf)}}}
\newcommand{\linkToPpt}[1]{\href{#1}{\blue{(ppt)}}}
\newcommand{\linkToCode}[1]{\href{#1}{\blue{(code)}}}
\newcommand{\linkToWeb}[1]{\href{#1}{\blue{(web)}}}
\newcommand{\linkToVideo}[1]{\href{#1}{\blue{(video)}}}
\newcommand{\award}[1]{\xspace} 

\newcommand{\myParagraph}[1]{{\bf #1.}}
\newcommand{\dcMRFc}{\scenario{DC-GM}}
\newcommand{\dcMRFd}{\scenario{DC-GMd}}

\newcommand{\rrr}{\scenario{RRR}}
\newcommand{\dcs}{\scenario{DCS}}
\newcommand{\DCGM}{\dcMRFc}

\newcommand{\PGO}{PGO\xspace}

\newcommand{\Od}{\text{O($d$)}}
\newcommand{\SOd}{\text{SO($d$)}}

\newcommand{\idiag}{\text{idiag}}
\newcommand{\newSEd}{(\SOd \times \Real{d})}
\newcommand{\constraints}{\substack{\vt_i \in \Real{d}  \\ \MR_i \in \SOd}}

\newcommand{\bmp}{\theta}
\newcommand{\Bmp}{\Theta}
\newcommand{\bmpt}{\bmp^t}
\newcommand{\bmpR}{\bmp^R}
\newcommand{\ldata}{\MQ}
\newcommand{\qdataOne}{\MU_{e}}
\newcommand{\qdataTwo}{\MW_{e}}

\newcommand{\noiset}{\vt^\epsilon}
\newcommand{\noiseR}{\MR^\epsilon}

\newcommand{\omegat}{\omega_t}
\newcommand{\omegaR}{\omega_r}

\newcommand{\tls}[1]{f_{#1}}
\newcommand{\barc}{\bar{c}}
\newcommand{\barct}{\barc_t}
\newcommand{\barcR}{\barc_R}
\newcommand{\tlst}{\tls{\barct}}
\newcommand{\tlsR}{\tls{\barcR}}

\newcommand{\barR}{\bar{\MR}}
\newcommand{\barT}{\bar{\MT}}
\newcommand{\bart}{\bar{\vt}}

\newcommand{\unarySet}{\calU}
\newcommand{\binarySet}{\calB}

\newcommand{\penaltyTerm}{\bar{c}}

\newcommand{\resRot}{\MR_j - \MR_{i} \barR_{ij}} 
\newcommand{\resTran}{\vt_j\!-\!\vt_i\!-\!\MR_{i} \bart_{ij}} 
\newcommand{\resPose}{\MT_j - \MT_i \barT_{ij} } 

\newcommand{\barcij}{\barc^{(ij)}_{(i'j')}}

\newcommand{\corSet}{\calC}

\newcommand{\nrLoops}{\ell}

\newcommand{\veccalG}{\calG} 

\newcommand{\modification}[1]{\textcolor{black}{#1}}

\newcommand{\finalmodification}[1]{\textcolor{black}{#1}}

\newcommand{\omitted}[2]{#2}
\newcommand{\shorten}[2]{#2}

\title{{Modeling Perceptual Aliasing in SLAM \\ via Discrete-Continuous Graphical Models}}

\author{Pierre-Yves Lajoie$^{1}$, Siyi Hu$^{2}$, 
Giovanni Beltrame$^{1}$, Luca Carlone$^{2}$
\thanks{Manuscript received: September 10, 2018; 
Revised: December 6, 2018; 
Accepted: January 6, 2019.
}
\thanks{This paper was recommended for publication by Editor Cyrill Stachniss upon evaluation of the Associate Editor and Reviewers' comments.} 
\thanks{This work was carried out during P.\,Lajoie's research stay in LIDS, and was partially funded by ARL DCIST CRA W911NF-17-2-0181, ONR RAIDER N00014-18-1-2828, and MIT Lincoln Laboratory.}
\thanks{$^{1}$P.\,Lajoie and G.\,Beltrame are with the Department of Computer and Software Engineering, \'Ecole Polytechnique de Montr\'eal, Montreal, Canada 
{\tt\footnotesize \{pierre-yves.lajoie,giovanni.beltrame\}@polymtl.ca}
}
\thanks{$^{2}$S.\,Hu and L.\,Carlone are with the Laboratory for 
Information \& Decision Systems (LIDS), Massachusetts Institute of Technology, Cambridge, USA, 
{\tt\footnotesize \{siyi,lcarlone\}@mit.edu}}
  \vspace{-0.5cm}
}
\begin{document}

\maketitle

\begin{abstract}
  Perceptual aliasing is one of the main causes of failure for Simultaneous
  Localization and Mapping (SLAM) systems operating in the wild.  Perceptual
  aliasing is the phenomenon where different places generate a similar visual
  (or, in general, perceptual) footprint.  This causes spurious measurements
  to be fed to the SLAM estimator, which typically results in incorrect
  localization and mapping results.  The problem is exacerbated by the fact
  that those outliers are \emph{highly correlated}, in the sense that
  perceptual aliasing creates a large number of mutually-consistent
  outliers. Another issue stems from the fact that \modification{most state-of-the-art techniques
   rely} on a \emph{given trajectory guess} (e.g., from odometry) to
  discern between inliers and outliers and this makes the resulting pipeline
  brittle, since the accumulation of error may result in incorrect choices and
  recovery from failures is far from trivial.  This work provides a 
  unified framework to \emph{model} perceptual aliasing in SLAM and provides
  practical algorithms that can cope with outliers 
  without relying on any initial guess.  We
  present two main contributions. The first is a \emph{Discrete-Continuous
    Graphical Model} (\DCGM) for SLAM: the continuous portion of the \DCGM captures the
  standard SLAM problem, while the discrete portion describes the
  selection of the outliers and models their correlation.  The second
  contribution is a semidefinite relaxation to perform
  inference in the \DCGM that returns estimates with provable sub-optimality guarantees.
  Experimental results on standard benchmarking datasets show that the proposed  
  technique compares favorably with state-of-the-art methods  
  \modification{while 
  not relying on an initial guess for optimization.}
\end{abstract}

\begin{tikzpicture}[overlay, remember picture]
\path (current page.north east) ++(-4,-0.2) node[below left] {
This paper has been accepted for publication in the IEEE Robotics and Automation Letters.
};
\end{tikzpicture}
\begin{tikzpicture}[overlay, remember picture]
\path (current page.north east) ++(-5.5,-0.6) node[below left] {
 Please cite the paper as: P. Lajoie, S. Hu, G. Beltrame and L. Carlone,
};
\end{tikzpicture}
\begin{tikzpicture}[overlay, remember picture]
\path (current page.north east) ++(-4.4,-1) node[below left] {
``Modeling Perceptual Aliasing in SLAM via Discrete-Continuous Graphical Models'',
};
\end{tikzpicture}
\begin{tikzpicture}[overlay, remember picture]
\path (current page.north east) ++(-7.1,-1.4) node[below left] {
  IEEE Robotics and Automation Letters (RA-L), 2019.
};
\end{tikzpicture}



\vspace{-0.3cm}


\section{Introduction}
\label{sec:intro}

\IEEEPARstart{S}{imultaneous} 
 Localization and Mapping (SLAM) is the backbone of several robotics applications.
SLAM is already widely adopted in consumer applications (e.g., robot vacuum cleaning, warehouse maintenance, virtual/augmented reality), 
and is a key enabler for truly autonomous systems operating in the wild, ranging from unmanned aerial vehicles operating in  
GPS-denied scenarios, to self-driving cars. 

Despite the \modification{remarkable advances} in SLAM, both researchers and practitioners are well aware of the brittleness of current SLAM systems. 
While SLAM failures are a tolerable price to pay in some consumer applications, they may put human life at risk in several safety-critical applications.
For this reason, SLAM is often avoided in those applications (e.g., self-driving cars) in favor of alternative solutions where the map
is built
beforehand in an offline (and typically human-supervised) manner, even though this implies extra setup costs. 

\begin{figure}[t]
\vspace{-0.6cm}
\hspace{0.7cm}
\subfloat{
  \includegraphics[trim=60mm 40mm 40mm 50mm, clip, width=0.8\linewidth]{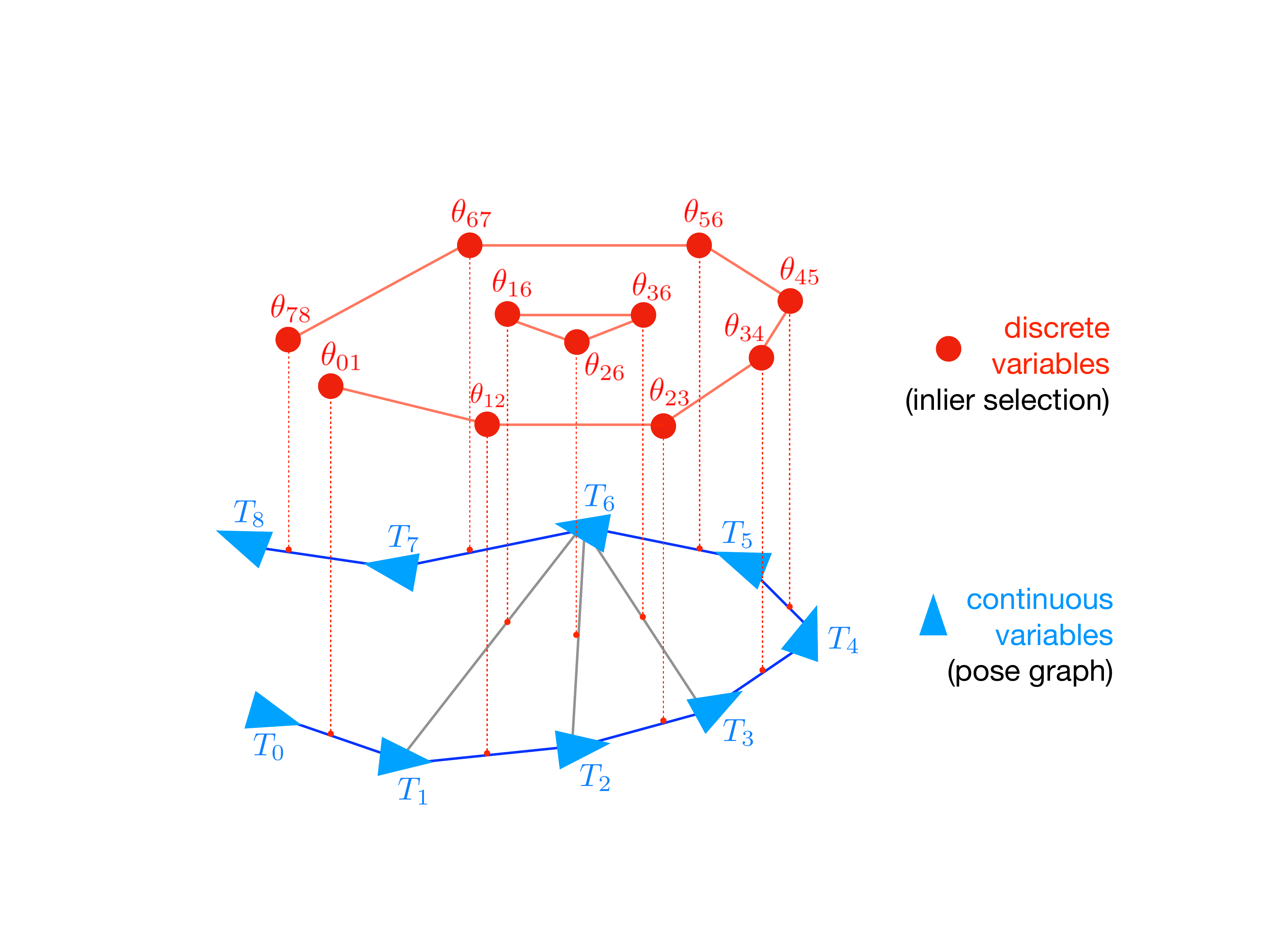}}  
\vspace{-3mm}
\caption{We introduce a \emph{Discrete-Continuous
    Graphical Model} (\DCGM) to model perceptual aliasing  in SLAM.
The model
    describes the interactions 
between continuous variables (e.g., robot poses) and discrete variables (e.g., the binary selection of 
inliers and outliers), and 
captures the correlation between the discrete variables
 (e.g., due to perceptual aliasing). 
\label{fig:dcMRF} }
\vspace{-6mm}
\end{figure} 

Arguably, the main cause of SLAM failure is the presence of incorrect data
association and outliers\shorten{~\cite{Cadena16tro-SLAMsurvey}.}{.} 
Incorrect data association is caused by \emph{perceptual aliasing}, the phenomenon 
where different places generate a similar visual (or, in general, perceptual) footprint. 
Perceptual aliasing leads to incorrectly associating the measurements taken by the robot to the wrong portion of the map, 
which \modification{may lead to map deformations and potentially to catastrophic failure of the mapping process. 
 The problem is exacerbated by the fact that those outliers are \emph{highly correlated}: 
due to the temporal nature of the data collection, 
 perceptual aliasing creates a large number of mutually-consistent outliers. 
 This correlation makes it even harder to judge if a measurement is an outlier, 
 contributing to the  brittleness of the resulting pipeline. 
 Surprisingly, while the SLAM literature extensively focused on mitigating the effects of perceptual aliasing, 
 none of the existing approaches attempt to explicitly model positive correlation between outliers.} 

 \myParagraph{Contribution} This work provides a unified framework to
 model perceptual aliasing and outlier correlation in SLAM. We propose \modification{a novel approach} to
 obtain provably-robust SLAM algorithms: rather than developing techniques to
 \emph{mitigate} the impact of perceptual aliasing, we \emph{explicitly model}
 perceptual aliasing using a \emph{discrete-continuous graphical model}
 (\DCGM).  A simple illustration is given in Fig.~\ref{fig:dcMRF}.
   The figure shows a \DCGM where the continuous variables, shown in
 blue, describe a standard SLAM formulation, i.e., a pose
 graph\shorten{~\cite{Cadena16tro-SLAMsurvey},}{,} where the triangles represent the
 trajectory of a moving robot while the edges represent measurements. The
 figure shows that we associate a discrete variable (large red circles) to
 each edge/measurement in the pose graph. The discrete variables decide
 between accepting or rejecting \modification{a measurement.}  The red edges in the top
 portion of the figure model the correlation between discrete variables.
The expert reader can recognize the top of the figure (graph in red), to be a discrete \emph{Markov Random Field} (MRF)~\cite{Blake11book-MRF}. 
The proposed model can naturally \modification{capture positive correlation between outliers}: for instance, we can model the correlation between three nearby edges, 
$(T_1,T_6)$, $(T_2,T_6)$, $(T_3,T_6)$ in Fig.~\ref{fig:dcMRF}, as a clique involving the corresponding \modification{discrete variables ($\theta_{16}$, $\theta_{26}$, $\theta_{36}$}) in the MRF (red triangle in the figure). Similarly, we can capture the temporal correlation of wheel slippage episodes by connecting variables corresponding to consecutive edges\omitted{. 
(e.g., $\theta_{12}$, $\theta_{23}$).}{.}

Our second contribution is the design of a semidefinite  (SDP)  relaxation that
computes a near-optimal estimate of the variables in the \DCGM.  Inference in
\DCGM is intractable in general, due to the nonconvexity of the corresponding
estimation problem and to the presence of discrete variables. 
We show how to obtain an SDP relaxation with per-instance sub-optimality guarantees, generalizing
previous work on provably-correct
SLAM without outliers~\cite{Carlone15icra-verification,Carlone16tro-duality2D,Carlone15iros-duality3D,Carlone18ral-robustPGO2D,Rosen16wafr-sesync}.
The SDP relaxation 
can be solved in polynomial time by off-the-shelf convex
solvers without relying on an initial guess. 

Our last contribution is an experimental evaluation on standard SLAM benchmarking datasets.
 The experimental results show that
 the proposed \DCGM model compares favorably with state-of-the-art methods,
 including 
\vertigo~\cite{Sunderhauf12iros}, \rrr~\cite{Latif12rss} and \dcs~\cite{Agarwal13icra}.
Moreover, they confirm that modeling  outlier correlation further increases the resilience of the proposed model, which is able to 
compute  
  correct SLAM estimates  even when $50\%$ of the loop closures are highly-correlated outliers. 
  Our current (Matlab) implementation is slow, compared to state-of-the-art methods, but 
  the proposed approach can be sped-up by designing a specialized solver along the lines of~\cite{Rosen16wafr-sesync}.
  We leave these numerical aspects (which are both interesting and non-trivial on their own) for future work.

\shorten{
\myParagraph{Paper structure} 
Section~\ref{sec:preliminaries} provides preliminary notions on MRFs
and pose graph optimization. Section~\ref{sec:DCGM} presents our new hybrid
discrete-continuous graphical model. Section~\ref{sec:convexRelax} presents a
semidefinite programming relaxation for inference in the \DCGM. 
Section~\ref{sec:experiments} presents experimental results, while Section~\ref{sec:conclusion} concludes the paper.
}{}



\section{Preliminaries and Related Work}
\label{sec:preliminaries}

This section reviews basic concepts about Markov Random Fields and Pose Graph Optimization. 


\subsection{Markov Random Fields (MRFs)}
\label{sec:MRF}

\emph{Markov Random Fields} (MRFs) are a popular graphical model for reconstruction and recognition problems in computer vision and robotics~\cite{Szeliski08pami-surveyMRF,Blake11book-MRF,Kappes15ijcv-energyMin}. 
A \emph{pairwise MRF} is defined by a set of $\ell$ \emph{nodes} we want to label, 
and a set of \emph{edges} or \emph{potentials}, representing probabilistic constraints involving the labels of a single or a pair of nodes.  
Here we consider \emph{binary} MRFs, where we associate a 
binary label 
 $\theta_i \in \{-1,+1\}$  to each node $i=1,\ldots,\ell$.

 The \emph{maximum a posteriori} (MAP) estimate of the variables in the MRF is
 the assignment of the node labels that attains the maximum of the posterior
 distribution of an MRF, or, equivalently, the minimum of the negative
 log-posterior~\cite{Szeliski08pami-surveyMRF}:
\modification{
\beq
\label{eq:MRF}
\min_{\substack{\theta_i \in \{-1,+1\} \\ i = 1,\ldots,\ell}}  \;\; 
-\sum_{i \in \unarySet} 
\penaltyTerm_{i} 
\theta_i
- \sum_{(i,j)\in\binarySet} 
\penaltyTerm_{ij} 
\theta_i \theta_j
\eeq
where $\unarySet \subseteq \{1,\ldots,\ell\}$ is the set of \emph{unary potentials} (terms involving a single node),
$\binarySet \subseteq \{1,\ldots,\ell\} \times \{1,\ldots,\ell\}$ is the set of \emph{binary potentials} (involving a pair of nodes).
Intuitively, if $\penaltyTerm_{i} > 0$ (resp. $\penaltyTerm_{i} < 0$), then the unary terms encourage $+1$ (resp. $-1$) labels for node $i$.  
Similarly, if $\penaltyTerm_{ij} > 0$, then the binary term $(i,j)$ encourages nodes $i$ and $j$ to have the same label (positive correlation) since that decreases the cost~\eqref{eq:MRF} by $\penaltyTerm_{ij}$.} 
While several choices of unary and binary potentials are possible, the 
expression in eq.~\eqref{eq:MRF} is a very popular model, and is referred to as the \emph{Ising model}~\cite[Section 1.4.1]{Blake11book-MRF}. 

Related works consider extensions of~\eqref{eq:MRF} to continuous~\cite{Fix14eccv}, discrete-continuous~\cite{Zach12eccv}, 
or discretized~\cite{Crandall12pami} labels, while to the best of our knowledge, 
our paper is the first to propose a semidefinite solver for discrete-continuous models  
and use these models to capture perceptual aliasing in SLAM.



\subsection{Pose Graph Optimization (\PGO)}
\label{sec:PGO}

Pose Graph Optimization (\PGO) is one of the most popular models for SLAM.
\PGO  consists in the estimation of a set of poses (i.e., rotations and translations) 
from pairwise relative pose measurements. 
 In computer vision a similar problem (typically involving only rotation) is used as a preprocessing step for bundle adjustment 
 in Structure from Motion (SfM)~\cite{Hartley13ijcv}.

\PGO estimates $n$ poses from $m$ relative pose measurements.
Each to-be-estimated pose $\MT_i \doteq [\MR_i \; \vt_i]$, $i=1,\ldots,n$,
 comprises a \emph{translation} vector $\vt_i \in \Real{d}$  
and a rotation matrix $\MR_i \in \SOd$, where $d=2$ in planar problems or $d=3$ in three-dimensional problems.
For a pair of poses $(i,j)$, 
a relative pose measurement $[\barR\ij \; \bart\ij]$, with $\bart\ij \in \Real{d}$ and 
$\barR\ij \in \SOd$,
 describes a noisy measurement of 
the relative pose between $\MT_i$ and $\MT_j$. 
Each measurement is assumed to be sampled from the following
generative  model:
\begin{align}
\label{eq:PGObinary}
\bart\ij = \MR_i\tran (\vt_j - \vt_i ) + \noiset\ij, \quad \quad \barR_{ij} = \MR_i\tran \MR_j \noiseR_{ij}
\end{align}
where $\noiset\ij \in \Real{d}$ 
and $\noiseR_{ij} \in \SOd$ represent translation and rotation measurement noise, respectively.
%
 PGO can be thought as an MRF\footnote{
 \modification{An interpretation of landmark-based SLAM as a pairwise MRF with continuous variables is  given by Dellaert in~\cite{Dellaert05rss}.} 
 } with variables living on manifold: 
we need to assign a pose to each node in a graph, given relative measurements associated to edges $\calE$ of the graph. 
The resulting graph 
is usually referred to as a \emph{pose graph}. 

Assuming the translation noise is Normally distributed 
with zero mean and information matrix $\omegat \eye_d$
and the rotation noise follows a Langevin distribution~\cite{Carlone15iros-duality3D,Carlone16tro-duality2D} with concentration parameter $\omegaR$, the MAP estimate for the unknown poses can be 
computed by solving the following optimization problem:
 \beq
 \displaystyle\min_{\constraints}  \!\!\displaystyle
 \!\!\!\!    \sum_{\;\;\;\;(i,j) \in \calE} \!\!\!\!\!\!
\omegat \|\resTran\|^2_2 
\;+\;  \frac{\omegaR}{2} \|\resRot\|^2_F
\label{eq:PGO}
\eeq
where $\|\cdot\|_F$ denotes the Frobenius norm. 
 The derivation of~\eqref{eq:PGO} is given  in~\cite[Proposition 1]{Carlone16tro-duality2D}.
The estimator~\eqref{eq:PGO} involves solving a nonconvex optimization, due to the nonconvexity of the set $\SOd$.
Recent results~\cite{Carlone16tro-duality2D,Rosen16wafr-sesync} show that 
one can still compute a globally-optimal solution to~\eqref{eq:PGO}, 
 when the measurement noise is reasonable, 
using convex relaxations. 

Unfortunately, the minimization~\eqref{eq:PGO} follows from the assumption that the measurement noise is light-tailed (e.g., Normally distributed translation noise) and it is known to produce completely wrong pose estimates when this assumption is violated, i.e., in presence of outlying measurements.

\begin{figure}[t]
\centering
\begin{minipage}{0.5\columnwidth} 
\centering
\includegraphics[width=\textwidth,trim=0mm 10mm 0mm 0mm,clip]{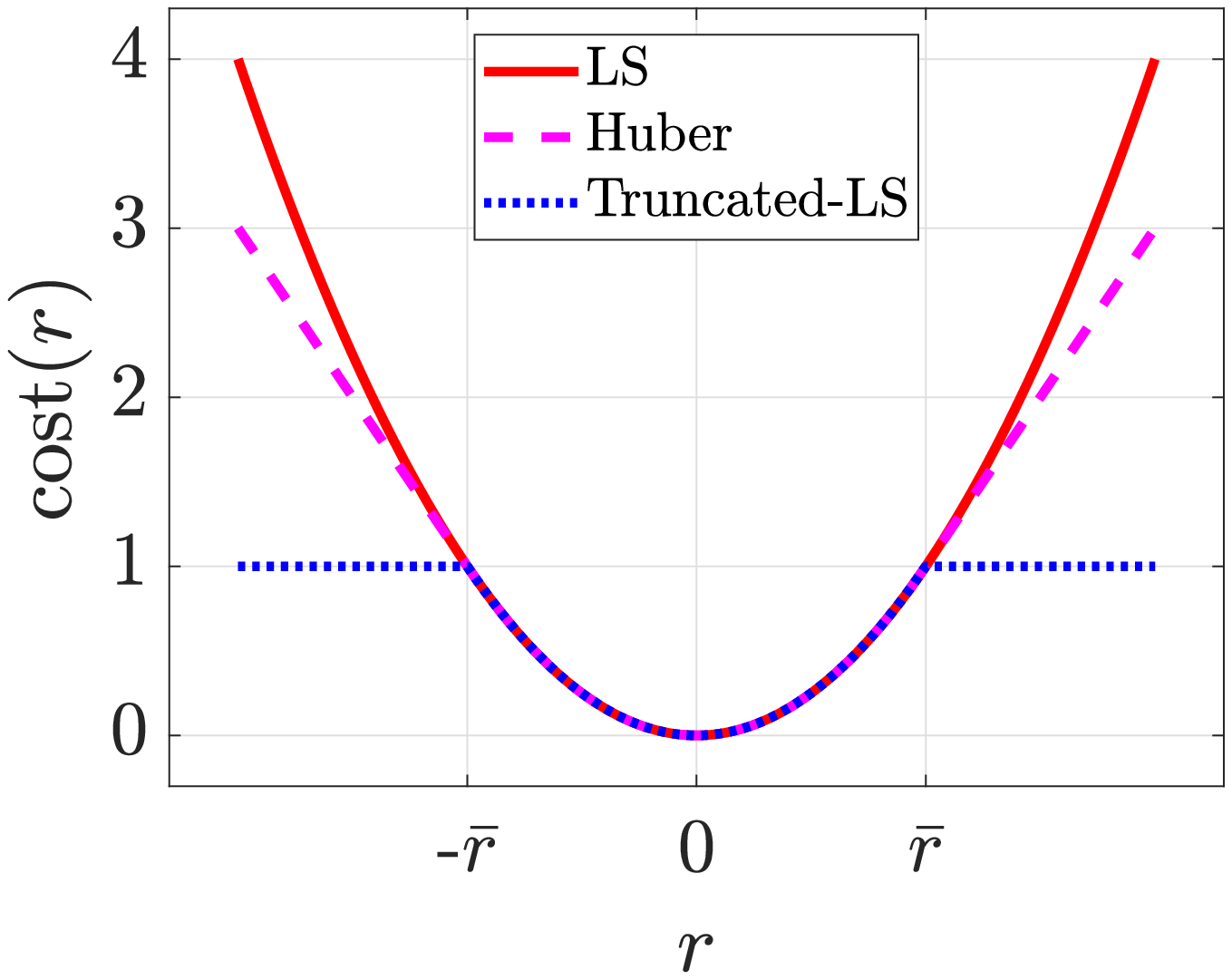}\vspace{-3mm}\\
\hspace{-3cm}(a)
\end{minipage}%
\begin{minipage}{0.5\columnwidth} 
\centering
\includegraphics[width=\textwidth,trim=0mm 10mm 0mm 0mm,clip]{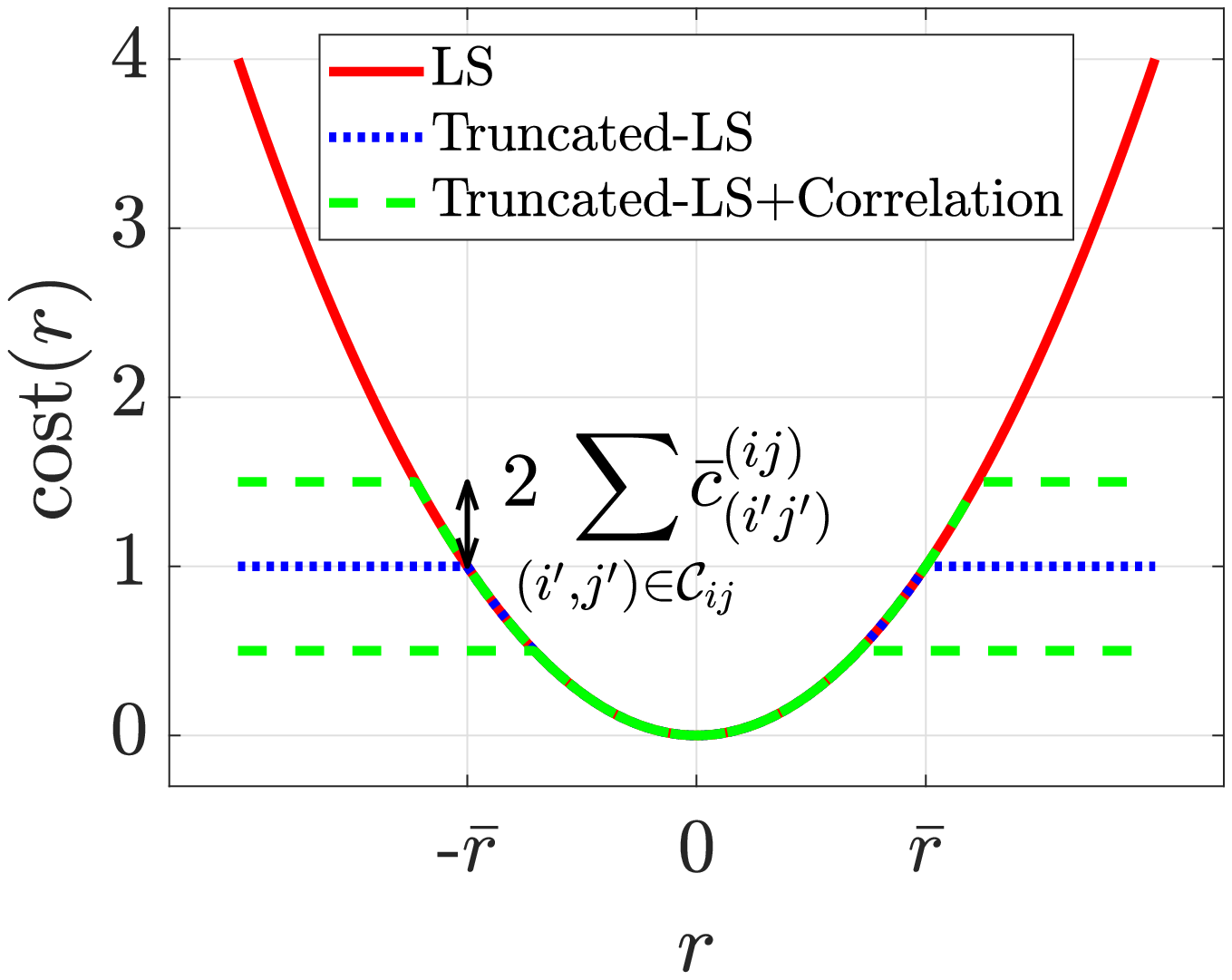}\vspace{-3mm}\\
\hspace{-3cm}(b)
\end{minipage}
\vspace{-3mm}
\caption{\finalmodification{(a) Cost associated to each residual error in the least squares (LS), Huber, and truncated LS 
 estimators. 
(b) The correlation terms $\barcij$ in eq.~\eqref{eq:robustPGOPA} have the effect of altering the error threshold in the truncated LS estimator.
}
\label{fig:robustCosts} }
\vspace{-6mm}
\end{figure} 

\subsection{Robust PGO}
\label{sec:robustPGO}

%
The sensitivity to outliers of the formulation~\eqref{eq:PGO} is due to the fact that we
 minimize the squares of the \emph{residual errors} (quantities appearing in the squared terms): 
 this implies that large residuals corresponding to spurious measurements dominate the cost. 
  Robust estimators reduce the impact of outliers by adopting cost functions that grow slowly (i.e., less than quadratically) 
  when the residual exceeds a given upper bound $\bar{r}$. This is the idea behind robust M-estimators, see~\cite{Huber81}.
  For instance, the Huber loss in Fig.~\ref{fig:robustCosts} grows linearly 
  outside the quadratic region $[-\bar{r},+\bar{r}]$. Ideally, one would like to adopt a truncated least squares (LS) formulation 
  (Fig.~\ref{fig:robustCosts}) where the impact of arbitrarily large outliers remains bounded. Such a formulation, 
  however, is non-convex and non-differentiable, typically making the resulting optimization hard.

Traditionally, outlier mitigation in SLAM and SfM relied on the 
use of robust M-estimators, see~\cite{Bosse17fnt,Hartley13ijcv}. 
Agarwal \etal~\cite{Agarwal13icra} propose \emph{Dynamic Covariance Scaling} (\dcs), which dynamically adjusts
 the measurement covariances to reduce the influence of outliers. 
Olson and Agarwal~\cite{Olson12rss} use  a max-mixture distribution to accommodate multiple 
hypotheses on the noise distribution of a measurement. 
Casafranca~\etal~\cite{Casafranca13iros} minimize the $\lone$-norm of the residual 
errors. 
Lee~\etal~\cite{Lee13iros} use expectation maximization. 
An alternative set of approaches attempts to explicitly \emph{identify} and reject outliers.
 \modification{Early techniques include RANSAC~\cite{Fischler81}
and branch \& bound~\cite{Neira01tra}. 
S\"{u}nderhauf and Protzel~\cite{Sunderhauf12iros,Sunderhauf12icra}
propose \vertigo, which augments the \PGO problem with latent binary variables (then relaxed to continuous variables) that are responsible for 
deactivating outliers. 
Latif \etal~\cite{Latif12rss}, Carlone \etal~\cite{Carlone14iros-robustPGO2D}, Graham \etal~\cite{Graham15iros}, 
 Mangelson \etal~\cite{Mangelson18icra} look for large sets of ``mutually consistent'' measurements. 
Pfingsthorn and Birk~\cite{Pfingsthorn13ijrr,Pfingsthorn16ijrr} 
 model ambiguous measurements using hyperedges and mixture of Gaussians, and provide a measurement selection approach that also constructs an initial guess for \PGO. 
 Both \cite{Olson12rss} and~\cite{Pfingsthorn16ijrr} implicitly model negative correlation 
 (or, more precisely, mutual exclusivity)
 between multiple edge hypotheses. 
 The introduction of discrete variables has also been used to reconcile data association and semantic  SLAM~\cite{Bowman17icra}, and to deal with unknown data association in SfM~\cite{Dellaert00cvpr}.}


\section{Discrete-continuous Graphical Models \\ for Robust Pose Graph Optimization}
\label{sec:DCGM}

We propose a novel approach for robust \PGO that addresses the three main limitations of the state of the art. 
\modification{First, rather than mitigating outlier correlation, we explicitly model it.}
 \modification{Second, our \PGO method (Section~\ref{sec:convexRelax}) does not rely on any initial guess.} 
 Third, we go beyond \modification{recently proposed} convex relaxations for 
 robust rotation and pose estimation~\cite{Wang13ima,Carlone18ral-robustPGO2D,Arrigoni18cviu}, and use a \emph{nonconvex} loss, namely, the \emph{truncated LS} cost in Fig.~\ref{fig:robustCosts}.
 This circumvents issues with convex robust loss functions
 which are known to have low breakdown point 
(e.g., the Huber loss~\cite{Carlone18ral-robustPGO2D} or $\ell_1$ norm~\cite{Wang13ima,Carlone18ral-robustPGO2D,Arrigoni18cviu} 
can be compromised by the presence of a single ``bad'' outlier).
%

\subsection{A unified view of robust \PGO}
 Let us partition the edges of the pose graph into odometric edges $\calE_{od}$ and loop-closure edges $\calE_{lc}$. 
  Perceptual aliasing affects exteroceptive sensors, hence ---while we can typically trust odometric edges--- 
  loop closures may include outliers. 

According to the discussion in Section~\ref{sec:robustPGO}, an ideal formulation for robust \PGO would use a truncated LS cost for the loop-closure edges in $\calE_{lc}$:
 %
 \beal
 \!\!\displaystyle\min_{\constraints}  \!\!\displaystyle
 \!\!   \sum_{\;\;\;\;(i,j) \in \calE_{od}} \!\!\!\!\!\!
 \omegat \|\resTran\|^2_2 
 \;+\;  \frac{\omegaR}{2} \|\resRot\|^2_F
 \label{eq:tlsPGO} \\
 + \!\!\!\!\displaystyle\sum_{(i,j) \in \calE_{lc}} \!\!\!\!
\omegat \tlst(\|\resTran\|_2) +  \frac{\omegaR}{2} \tlsR(\|\resRot\|_F) 
\eeal
where, for a positive scalar $c$, the function $\tls{c}(\cdot)$ is: 

 \beq
\tls{c}(x) = \left\{
\begin{array}{ll}
x^2 & \text{if } |x| \leq c \\
 c^2 & \text{otherwise}
\end{array}	
\right.
\label{eq:tls}
\eeq

While the formulation~\eqref{eq:tlsPGO} would be able to tolerate arbitrarily ``bad'' outliers, 
it has two main drawbacks. First, $\tls{c}(\cdot)$ is non-convex, adding to the non-convexity already induced by the rotations 
(\SO{d} is a non-convex set). Second, the cost is non-differentiable, as shown in~Fig.~\ref{fig:robustCosts}, 
hence also preventing the use of fast (but local) smooth optimization techniques.

The first insight behind the proposed approach is simple but powerful: 
we can rewrite the truncated LS cost~\eqref{eq:tls} as a minimization over a binary variable: 
 \beq
\tls{c}(x) = \min_{\theta \in \{-1;+1\}} \frac{(1+\theta)}{2} x^2 + \frac{(1-\theta)}{2} c^2 
\label{eq:tls2}
\eeq
To show the equivalence between~\eqref{eq:tls2} and~\eqref{eq:tls}, we observe that for any $\bar{x}$ such that $\bar{x}^2 < c^2$ (or $|\bar{x}| < c$), 
the minimum in~\eqref{eq:tls2} is attained for $\theta=+1$ and $\tls{c}(\bar{x}) = \bar{x}^2$; on the other hand, for any $\hat{x}$ such that
$\hat{x}^2 > c^2$ (or $|\hat{x}| > c$), the minimum in~\eqref{eq:tls2} is attained for $\theta=-1$ and $\tls{c}(\hat{x}) = c^2$. 

We can now use the expression~\eqref{eq:tls2} to rewrite the cost function~\eqref{eq:tlsPGO} by introducing a binary variable 
for each rotation and translation measurement: 
%
 \beal
 \vspace{-6mm}
 \!\!\!\!\!\!\displaystyle\min_{ \substack{\vt_i \in \Real{d}  \\ \MR_i \in \SOd \\ \bmpt_{ij} \in \{-1;+1\} \\ \bmpR_{ij} \in \{-1;+1\}  } }  
\!\!& \!\!\!\!\displaystyle\!\!\!\!  \!\! \sum_{\;\;\;\;(i,j) \in \calE_{od}} \!\!\!\!\!\!
 \omegat \|\resTran\|^2_2 
 \;+\;  \frac{\omegaR}{2} \|\resRot\|^2_F +
 \nonumber
 \eeal
  \beal
&   \!\!\!\!\!\!\displaystyle +\!\!\!\!\!\!\!\!\sum_{\;\;\;\;(i,j) \in \calE_{lc}} \!\!\!\!\!\!
 \omegat \frac{(1+\bmpt_{ij})}{2} \|\resTran\|_2^2 + \omegat \frac{(1-\bmpt_{ij})}{2} \barct^2
\\ 
& \displaystyle  +  \frac{\omegaR}{2} \frac{(1+\bmpR_{ij})}{2} \|\resRot\|_F^2 + \frac{\omegaR}{2} \frac{(1-\bmpR_{ij})}{2} \barcR^2
\label{eq:robustPGOmp}
\eeal
where $\barct$ and $\barcR$ are simply the largest admissible residual errors for a translation and rotation measurement
to be considered an inlier. Intuitively, $\bmpt_{ij}$ decides whether a translation measurement is an inlier ($\bmpt_{ij}=+1$) 
or an outlier ($\bmpt_{ij}=-1$); $\bmpR_{ij}$ has the same role for rotation measurements. 
While eq.~\eqref{eq:robustPGOmp} resembles formulations in the literature, e.g., S\"{u}nderhauf's switchable constraints~\cite{Sunderhauf12iros}, 
\modification{establishing connections with the truncated LS cost provides a physically meaningful 
interpretation of the parameters $\barct$ and $\barcR$ (maximum admissible residuals).} 
 Moreover, we will push the boundary of the state of the art by modeling the outlier correlation (next sub-section) and 
 proposing global semidefinite solvers (Section~\ref{sec:convexRelax}).

\subsection{Modeling outlier correlation and perceptual aliasing}


The goal of this section is to introduce extra terms in the cost~\eqref{eq:robustPGOmp} to model the correlation between 
subsets of binary variables, hence capturing outlier correlation. 
\modification{For the sake of simplicity}, we assume that a unique binary variable is used to decide if both the translation 
and the rotation components of measurement $(i,j)$ are accepted, i.e., we set $\bmpt_{ij}= \bmpR_{ij} \doteq \bmp_{ij}$. 
This assumption is not necessary for the following derivation, 
 but it allows using a more compact notation.
In particular, we rewrite~\eqref{eq:robustPGOmp} more succinctly as:
 \beal
 \!\!\displaystyle\min_{ \substack{  \MT_i \in \SOd \times \Real{d}  \\ \bmp_{ij} \in \{-1;+1\} } } 
&\displaystyle \!\!\!\! \sum_{\;\;\;\;(i,j) \in \calE_{od}} \!\!\!\!\!\!
  \| \resPose \|_\MOmega^2\\ 
&\!\!\!\!\!\!\!\!\!\!\!\!\finalmodification{  + \!\!\!\!\displaystyle \!\!\!\! \sum_{\;\;\;\;(i,j) \in \calE_{lc}} \!\!\!\!\!\!
\frac{(1+\bmp_{ij})}{2} \| \resPose \|_\MOmega^2 + \frac{(1-\bmp_{ij})}{2} \barc}
\label{eq:robustPGOT}
\eeal
where for two matrices $\MM$ and $\MOmega$ of compatible dimensions $\|\MM\|^2_\MOmega \doteq \trace{\MM \MOmega \MM \tran}$,
and --following~\cite{Briales17ral}-- we defined:
\beq
\MT_i \doteq [\MR_i \; \vt_i], 
\quad 
\barT_{ij} \doteq \matTwo{\barR_{ij} & \bart_{ij} \\ \zero_d\tran & 1},
\quad
\MOmega \doteq \matTwo{\frac{\omegaR}{2} \eye_d & \zero_d \\ \zero_d\tran & \omegat} 
\nonumber
\eeq
and for simplicity we called $\barc \doteq \omegaR \barct^2 +  \frac{\omegat}{2} \barcR^2$.

We already observed in Section~\ref{sec:MRF} that to model 
the correlation between two discrete  variables $\bmp_{ij}$ and $\bmp_{i'j'}$ \modification{we can add terms $-\barcij \bmp_{ij} \bmp_{i'j'}$
to the cost function, which penalize a mismatch between $\bmp_{ij}$ and $\bmp_{i'j'}$
 whenever the scalar $\barcij$ is positive.} 
This leads to generalizing problem~\eqref{eq:robustPGOT} as follows:
 \beal
 \!\!\displaystyle\min_{ \substack{  \MT_i \in \SOd \times \Real{d}  \\ \bmp_{ij} \in \{-1;+1\} } }  
&\displaystyle \!\!\!\! \sum_{\;\;\;\;(i,j) \in \calE_{od}} \!\!\!\!\!\!
\| \resPose \|_\MOmega^2  \label{eq:robustPGOPA}\\ 
&  \!\!\!\!\!\!\!\!\!\!\!\!\!\!\!\!\!\!\!\!\!\!\!\!\!\!\!\!\!\!\!\!\!\!
\finalmodification{+ \!\!\!\!\!\!\displaystyle \!\!\!\! \sum_{\;\;\;\;(i,j) \in \calE_{lc}} \!\!\!\!\!\!\!\!\!
\frac{(1+\bmp_{ij})}{2} \| \resPose \|_\MOmega^2 + \frac{(1-\bmp_{ij})}{2} \barc
- \!\!\!\!\!\!\!\! \!\!\!\!\!\! \displaystyle\sum_{\;\;\;\;(i,j), (i',j') \in \corSet}\!\!\!\!\!\!\!\!\!\!\!\!\! \barcij \bmp_{ij} \bmp_{i'j'}}
\eeal
where the set $\corSet$ contains pairs of edges that are correlated, i.e., pairs of edges $(i,j), (i',j')$ for which 
if $(i,j)$ is found to be an outlier, it is likely for $(i',j')$ to be an outlier as well. 

\finalmodification{In the supplemental material~\cite{Lajoie18tr-DCGM}, we show that the correlation terms have the effect of altering the threshold $\barc$. 
For instance, if all neighbors $(i',j')$ of an edge $(i,j)$ are inliers ($\bmp_{i'j'}=1$), the correlation terms become $\barcij \bmp_{ij}$ and they have the effect of increasing $\barc$. We also show that
perturbations of $\barc$ are bounded in the interval $[ \barc - 2\sum_{(i',j') \in \calC_{ij}} \!\barcij, \barc + 2\sum_{(i',j') \in \calC_{ij}} \!\barcij]$, where $\corSet_{ij}$ is the set of edges correlated to the edge $(i,j)$, see Fig.\ref{fig:robustCosts}(b) for an illustration.
}

Problem~\eqref{eq:robustPGOPA} describes a \emph{discrete-continuous graphical model} (\DCGM) as the one pictured 
in Fig.~\ref{fig:dcMRF}: the optimization problems returns the most likely assignment of variables in the graphical 
model, which contains both continuous variables ($\MT_i$) and discrete variables ($\bmp_{ij}$). 
The reader can notice that if the assignment of discrete variables is given,~\eqref{eq:robustPGOPA} 
reduces to \PGO, while if the continuous variables are given, then~\eqref{eq:robustPGOPA} becomes an MRF, 
where the second sum in~\eqref{eq:robustPGOPA} defines the unary potentials for each discrete variable in the MRF.


\section{Inference in \DCGM via Convex Relaxation}
\label{sec:convexRelax}

The \DCGM presented in Section~\ref{sec:DCGM} captures two very desirable aspects: 
(i) it uses a robust truncated LS loss function 
and (ii) it can easily model outlier correlation. On the downside, the optimization~\eqref{eq:robustPGOPA} 
is intractable in general, due to the presence of discrete variables and the non-convex nature of 
the rotation set $\SOd$. 

Here we derive a convex relaxation that is able to 
compute near-optimal solutions for~\eqref{eq:robustPGOPA} in polynomial time. 
While we  do not expect to compute exact solutions for~\eqref{eq:robustPGOPA} in all cases in polynomial time 
(the problem is NP-hard in general), 
our goal is to obtain a relaxation that works well when the noise on the inliers 
is reasonable (i.e., similar to the one found in practical applications) and whose quality is largely insensitive to the 
presence of a large number of (arbitrarily ``bad'') outliers.

In order to derive our convex relaxation, it is convenient to reformulate~\eqref{eq:robustPGOPA} using a more compact matrix notation.
 Let us first 
 ``move'' the binary variables inside the norm and drop constant terms from the objective in~\eqref{eq:robustPGOPA}:
%
%
 \beal
 \hspace{-0.4cm}
\displaystyle\min_{ \substack{  \MT_i \in \SOd \times \Real{d}  \\ \bmp_{ij} \in \{-1;+1\} } }  
& \displaystyle \hspace{-0.7cm}\sum_{\;\;\;\;(i,j) \in \calE_{od}} \hspace{-0.4cm} \| \resPose \|_\MOmega^2  
+ \!\!\!\!\displaystyle \!\!\!\! \sum_{\;\;\;\;(i,j) \in \calE_{lc}} 
\hspace{-0.5cm}\modification{\| \frac{(1+\bmp_{ij})}{2}(\resPose) \|_\MOmega^2 }  \\
& \displaystyle - \sum_{(i,j) \in \calE_{lc}}  \frac{\bmp_{ij}}{2} \barc  \;\;\modification{- \!\!\!\! \sum_{\;\;\;\;(i,j), (i',j') \in \calC}  \barcij  \bmp_{ij} \bmp_{i'j'}}
\label{eq:robustPGOsplit}
\eeal
\modification{where we noted that $\frac{(1+\bmp_{ij})}{2}$ is either zero or one, hence it can be safely moved inside the norm, 
and  we dropped $\frac{1}{2} \barc$.} 

We can now stack pose variables into a single $d \times (d+1)n$ matrix $\MT \doteq [\MT_1 \, \ldots \, \MT_n]$. 
We also use a \emph{matrix representation} for the binary variables 
$\Bmp \doteq [\Bmp_1 \, \ldots \, \Bmp_\nrLoops] \in \{-\MI_d; +\MI_d\}^\nrLoops$ where $\nrLoops = |\calE_{lc}|$ denotes the number of loop closures 
and $\MI_d$ denotes the identity matrix of size $d$. Finally, we define:
\bea
	\label{eq:X}
	\MX = [\MT\quad\Bmp\quad \MI_d] \in \newSEd^n\times\{-\MI_d; +\MI_d\}^\nrLoops\times \MI_d 
	\nonumber\\
	\left(\text{note: } \MX\tran \MX = 
	\begin{bmatrix}
		\MT\tran \MT & \MT\tran\Bmp & \MT\tran \\
		\Bmp\tran \MT & \Bmp\tran\Bmp & \Bmp\tran \\
		\MT & \Bmp & \MI_d 
	\end{bmatrix}\right)
\eea

The following proposition provides a compact reformulation of problem~\eqref{eq:robustPGOsplit} using the matrix $\MX$ in~\eqref{eq:X}: 

\begin{proposition}[Inference in \DCGM]
\label{prop:matrix}
Problem \eqref{eq:robustPGOsplit} can be equivalently written in compact form using the matrix variable $\MX$ in~\eqref{eq:X} 
as follows:
\begin{equation}
\label{eq:matrixProblem}
\begin{array}{rl}
\!\!\displaystyle\min_{\MX  }  
\!\!\displaystyle &\trace{\ldata\MX\tran\MX}  
+ \!\!\!\!\displaystyle \!\!\!\!\sum_{\;\;e=(i,j) \in \calE_{lc}} \!\!\!\!
\trace{\qdataOne\MX\tran\MX\qdataTwo\MX\tran\MX} \\
\subject & \MX\in \newSEd^n \times\{-\MI_d; +\MI_d\}^\nrLoops \times \MI_d
\end{array}
\end{equation}
where $\ldata, \qdataOne, \qdataTwo\in \Real{ \left(n(d+1)+d\nrLoops+d\right) \times \left(n(d+1)+d\nrLoops+d\right)}$ 
are sparse matrices (for all loop closures $e \in \calE_{lc}$). The expressions for these (known) matrices \modification{are} given in Appendix.  
\end{proposition}

Intuitively, $\ldata$ in~\eqref{eq:matrixProblem} captures \modification{the terms in the first, third, and fourth sum in~\eqref{eq:robustPGOsplit}, while 
the 
 sum including $\qdataOne, \qdataTwo$ (one term for each loop closure $e$) captures the terms in the second sum in~\eqref{eq:robustPGOsplit} 
which couples discrete and continuous variables.}

\modification{The final step before obtaining a convex relaxation is to write the ``geometric'' constraints in~\eqref{eq:matrixProblem}
in terms of linear algebra. Towards this goal, we relax the set $\SOd$ (rotation matrices) to $\Od$ (orthogonal matrices), i.e., 
we drop the constraint that rotation matrices need to have determinant $+1$.
 In related work, we found the determinant constraint to be redundant~\cite{Tron15rssws3D-dualityPGO3D}. 
 Moreover, this is done for the sake of simplicity, while the determinant constraints can be still modeled as shown in~\cite{Tron15rssws3D-dualityPGO3D}. 
Then, we obtain an SDP relaxation of Problem~\eqref{eq:matrixProblem} by (i) introducing a matrix variable $\MZ = \MX\tran\MX$ and rewriting~\eqref{eq:matrixProblem} as a function of $\MZ$, 
(ii) noting that any matrix $\MZ = \MX\tran\MX$ is a positive-semidefinite ($\MZ \succeq 0$) rank-d matrix, and (iii) relaxing the non-convex rank-d constraint.
}
\begin{proposition}[Semidefinite Relaxation of \DCGM] The following SDP is a convex relaxation of Problem~\eqref{eq:matrixProblem}:
\begin{equation}
\label{eq:SDP}
\hspace{-2mm}
\begin{array}{rll}
\displaystyle\min_{\MZ}   &\multicolumn{2}{l}{\trace{\ldata\MZ} + \!\!\!\!\displaystyle \!\!\!\! \sum_{e=(i,j) \in \calE_{lc}}  \!\!\!\! \trace{\qdataOne\MZ\qdataTwo\MZ}}\\
\subject & [\MZ]_{ii} = \matTwo{\MI_d & * \\ * & *} & \substack{i=1,\ldots,n}\\
& [\MZ]_{ii} = \MI_d   &  \substack{i=n+1,\ldots,n+\ell+1}\\
& [\MZ]_{ij} = \emph{\idiag}([\MZ]_{ij})  &  \substack{i,j=n+1,\ldots,n+\ell+1}\\
		 & \MZ \succeq 0
\end{array}
\end{equation}
\modification{where $[\MZ]_{ij}$ denotes the block of $[\MZ]$ in block row $i$ and block column $j$, 
 the symbol ``*'' denotes entries that are unconstrained (we follow the notation of~\cite{Briales17ral}), 
and where $[\MZ]_{ij} = \emph{\idiag}([\MZ]_{ij})$ enforces the block $[\MZ]_{ij}$ 
to be an isotropic diagonal matrix, i.e., a scalar multiple of $\eye_d$.}
\end{proposition}
\modification{
Let us explain the constraints in~\eqref{eq:SDP}, by using the block structure of $\MZ$ described in~\eqref{eq:X}.
 For $i=1,\ldots,n$, the diagonal blocks $[\MZ]_{ii}$ are in the form of $\MT_i\tran \MT_i$, hence 
 the first constraint in~\eqref{eq:SDP} captures the orthogonality of the rotation matrix included in each pose $\MT_i$. 
 For $i=n+1,\ldots,n+\ell$, the diagonal blocks $[\MZ]_{ii}$ are in the form of $\Bmp_i\tran \Bmp_i$ and since $\Bmp_i \in \{-\eye_d,+\eye_d\}$, 
$\Bmp_i\tran \Bmp_i = \eye_d$, which is captured in the second constraint in~\eqref{eq:SDP}; 
similar considerations hold for $i = \nrLoops+1$.
 Finally, the products $\Bmp_i\tran \Bmp_j$ (captured by the blocks $[\MZ]_{ij}$ when $i,j = n+1,\ldots,n+\ell+1$) 
  must be diagonal matrices, producing the third constraint in~\eqref{eq:SDP}. }

The SDP relaxation can be solved using off-the-shelf convex solvers. 
In particular, we note that the constraint $[\MZ]_{ij} = {\idiag}([\MZ]_{ij})$ can be implemented as a set of linear equality constraints. \modification{Indeed, this constraint can be rewritten as $[\MZ]_{ij} = 
[\MZ]_{ij,11} \cdot \eye_d$ where $[\MZ]_{ij,11}$ is the top left entry of $[\MZ]_{ij}$. Therefore, the constraint enforces that the matrix has offdiagonal elements equal to zero and diagonal elements equal to a single scalar $[\MZ]_{ij,11}$.}
The SDP relaxation~\eqref{eq:SDP}
 enjoys the typical per-instance optimality guarantees described in related work~\cite{Carlone16tro-duality2D,Carlone15iros-duality3D,Carlone18ral-robustPGO2D,Rosen16wafr-sesync}.
In particular, if the solution $\MZ^\star$ of~\eqref{eq:SDP} has rank $d$, then the relaxation solves~\eqref{eq:matrixProblem} \emph{exactly}.
Moreover, the optimal objective of~\eqref{eq:SDP} is a lower bound for the optimal objective~\eqref{eq:matrixProblem}, a property that can be 
used to evaluate how sub-optimal a given estimate is, see~\cite{Carlone16tro-duality2D,Carlone15iros-duality3D}.  


	


\section{Experiments}
\label{sec:experiments}

This section presents two sets of experiments. 
Section~\ref{sec:simulations} reports the results of Monte Carlo runs on a synthetic dataset 
and shows that the proposed technique compares favorably with the state of the art, and that 
 modeling outlier correlation leads to performance improvements.
Section~\ref{sec:realExperiments} evaluates the proposed techniques in three real benchmarking datasets 
and shows that our approach outperforms related techniques \modification{while not requiring any initial guess.} 

\subsection{Experiments On Synthetic Dataset}
\label{sec:simulations}

\begin{figure}
 \vspace{-3mm}
\centering
\subfloat[\dcMRFc]{
  \includegraphics[width=0.4\linewidth]{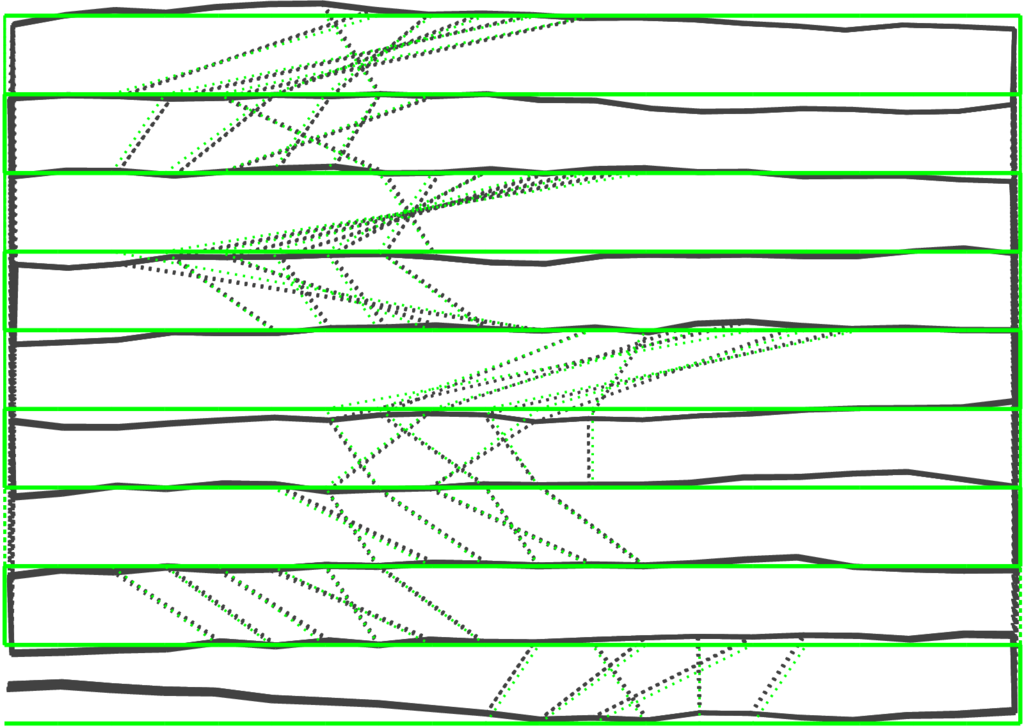} }  
\subfloat[\dcMRFd]{
  \includegraphics[width=0.4\linewidth]{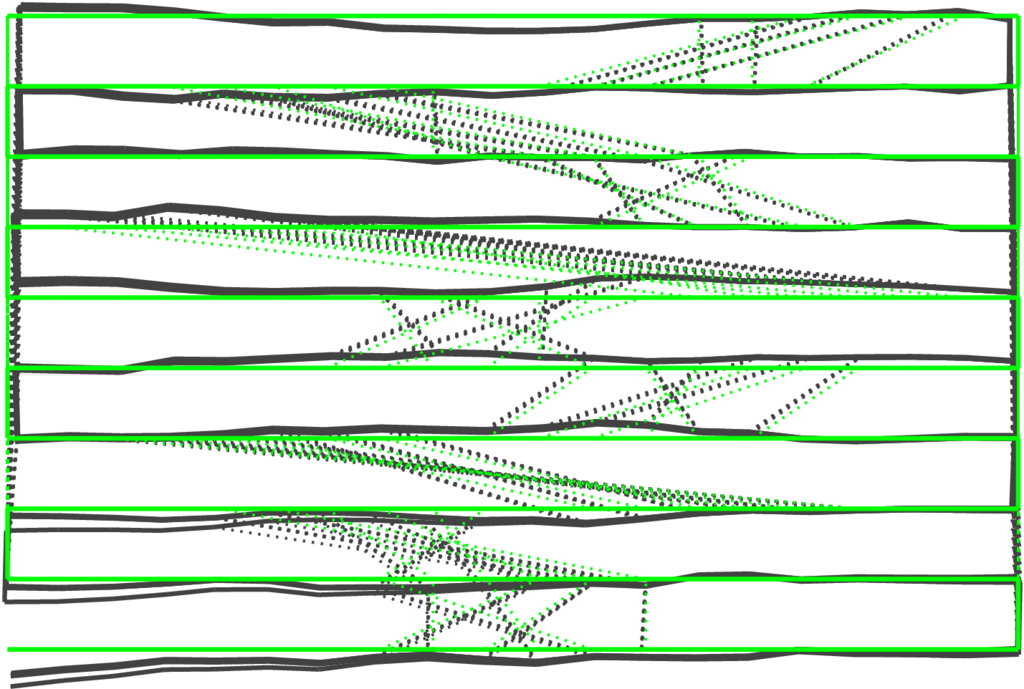} }    
   \vspace{-1mm}       
\caption{Trajectory estimates computed by the proposed techniques (black, overlay of 5 runs) versus ground truth (green)  
for the simulated grid dataset.
 }
\label{fig:syntheticShadowPlots} \vspace{-5mm}
\end{figure}

\myParagraph{Methodology}
For this set of experiments, we built a synthetic dataset composed of a simple trajectory on a grid of 20 by 10 nodes.
Then we added random groups of loop closures between the rows as described in \cite{Sunderhauf12iros}. 
Typically, in presence of perceptual aliasing, the outliers are in mutually-consistent groups, e.g., the SLAM front-end generates multiple false loop 
closures in sequence. 
To simulate this phenomenon, we set the loop closures in each group to be either all inliers or all outliers.
We set the standard deviation of the translation and rotation noise for the inlier measurements (odometry and correct loop closures) 
to $0.1$m and $0.01$rad. 
The maximum admissible errors for the truncated LS~\eqref{eq:tls} is set to $1\sigma$ of the measurement noise. 
We tested the performance of our techniques for increasing levels of outliers, up to the case where $50\%$ of the loop closure are outliers. 
Fig.~\ref{fig:syntheticShadowPlots} shows the overlay of multiple trajectories (5 runs) estimated by our techniques 
versus the ground truth trajectory (green),  when $50\%$ of the loop closures are outliers.

\myParagraph{Compared Techniques}
We evaluate the performance of 
the proposed technique, \dcMRFc, which solves the minimization problem
\eqref{eq:robustPGOPA}. 
In order to show that capturing outlier correlation leads to performance improvements, 
we also test a variation of the proposed approach, 
called \dcMRFd, which implements the minimization problem
\eqref{eq:robustPGOT}, where outliers are assumed uncorrelated (the ``d'' stands for decoupled). 
In both \dcMRFc and \dcMRFd, we solve the SDP using \cvx~\cite{CVXwebsite} in Matlab. 
If the resulting matrix does not have rank $d=2$ (in which case we are not guaranteed to get an exact solution 
to the non-relaxed problem),
  we round the result to detect the set of outliers,  and re-run the optimization without the outliers.

We benchmarked our approach against three other robust PGO techniques, i.e., \vertigo~\cite{Sunderhauf12iros}, \rrr~\cite{Latif12rss} and \dcs~\cite{Agarwal13icra}. For  \vertigo we use the default parameters, while for \rrr and \dcs we report results for multiple choices of parameters, since these parameters have a significant impact on  performance.
In particular, for \rrr we consider three cluster sizes ($t_g = \{1, 5, 10\}$) 
and for \dcs we considered three values of the parameter $\Phi = \{1, 10, 100\}$~\cite{Agarwal13icra}. 
For all these techniques, we used the odometric estimate as initial guess.

\begin{figure}
\centering
\subfloat{
  \includegraphics[width=0.6\linewidth]{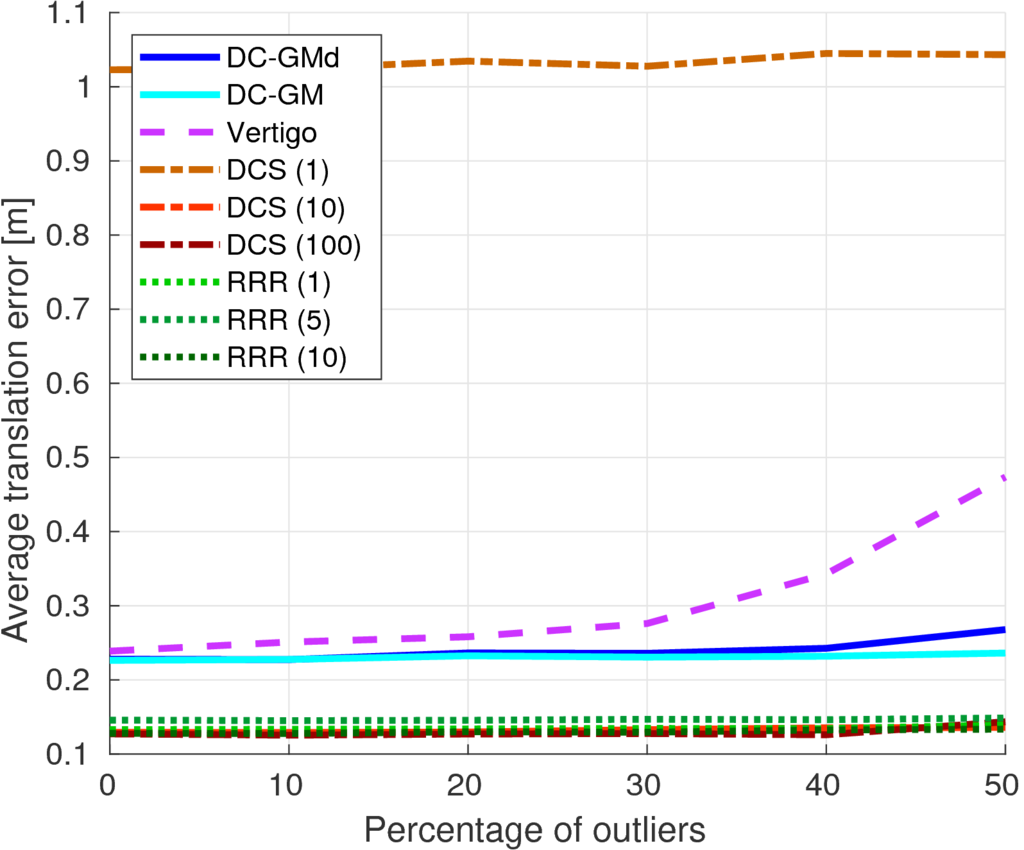}} 
  \vspace{-3mm}
\caption{Average translation error of the 9 approaches tested in this paper with an increasing percentage of outliers.}
\label{fig:syntheticErrorPlots}
\end{figure}

\myParagraph{Results and Interpretation}
Fig.~\ref{fig:syntheticErrorPlots} reports the average translation error 
for all the compared approaches and for 
increasing percentage of outliers.
\vertigo's error grows quickly beyond 30\% of outliers. 
For \dcs, the performance heavily relies on correct parameter tuning: for some choice of parameters 
($\Phi=\{10,100\}$) it has excellent performance while the approach fails for $\Phi=1$. Unfortunately, these parameters are difficult to tune in general
 (we will observe in Section~\ref{sec:realExperiments} that the choice of parameters mentioned above may not produce the best results in the real tests). 
The proposed techniques, \dcMRFd and \dcMRFc, compare favorably against the state of the art while they are slightly less accurate than 
\rrr, which produced the best results in simulation.

\begin{figure}
\centering
\subfloat{
	\includegraphics[width=0.7\linewidth]{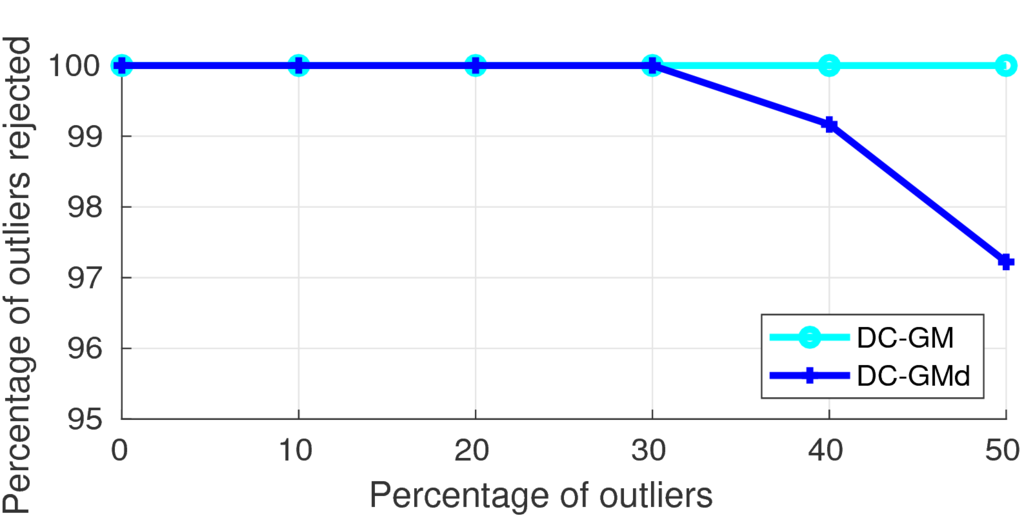} } 
\caption{Percentage of rejected outliers for the proposed techniques.}\vspace{-5mm}
\label{fig:syntheticInOutPlots}
\end{figure}

In order to shed light on the performance of \dcMRFc and \dcMRFd,
Fig.~\ref{fig:syntheticInOutPlots} reports the average percentage of outliers rejected by these two techniques. 
While from the scale of the y-axis we note that both techniques are able to reject most outliers, 
\dcMRFc is able to reject \emph{all} outliers in all tests even when up to $50\%$ of the loop closures are spurious.
As expected, modeling outlier correlation as in \dcMRFc improves outlier rejection performance.  
We also recorded the number of \emph{incorrectly rejected inliers}: both approaches do not reject any inlier 
and for this reason we omit the corresponding figure. 

\modification{In our tests, 
the SDP relaxation~\eqref{eq:SDP} typically produces low-rank solutions with 2 relatively large eigenvalues, 
followed by 2 smaller ones (the remaining eigenvalues are numerically zero).
The interested reader can find statistics on the average rank, results for different choices of the thresholds $\barc$ and $\barcij$, and additional tests in a simulated Manhattan World in the supplemental material~\cite{Lajoie18tr-DCGM}.
}

\begin{figure}
\centering
\hspace{-5mm}
\subfloat[CSAIL]{
  \includegraphics[width=0.33\linewidth]{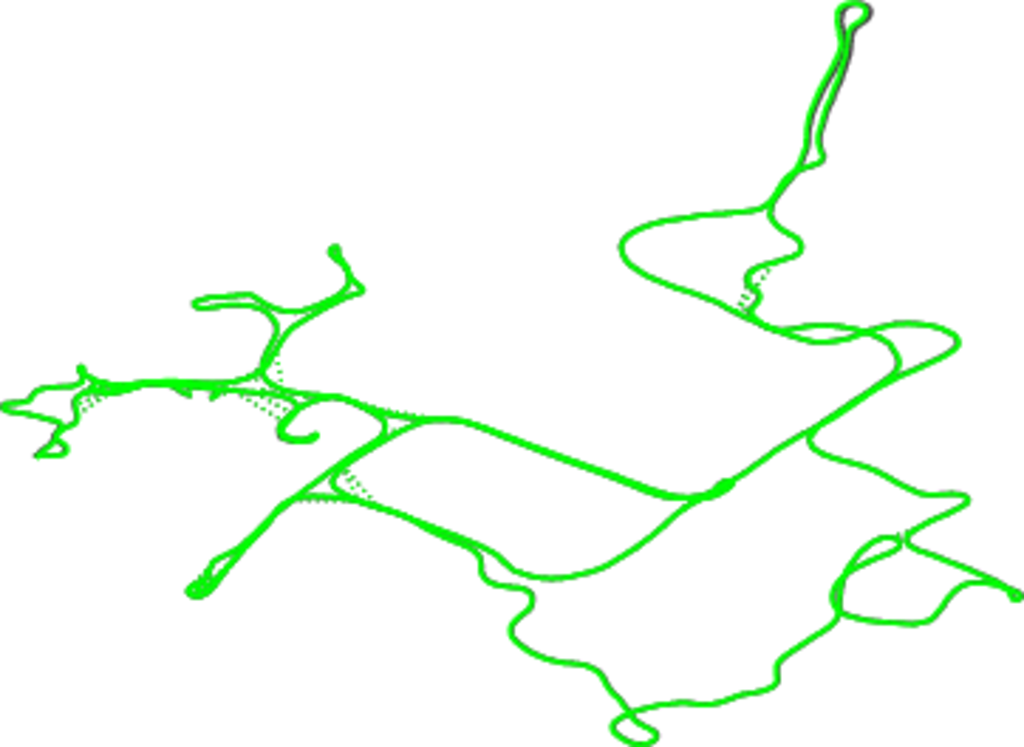} } 
\subfloat[FR079]{
  \includegraphics[width=0.33\linewidth]{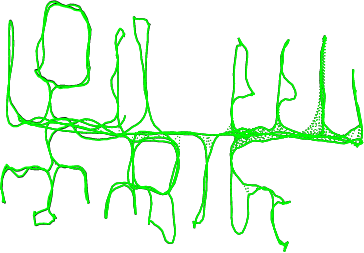}} 
\subfloat[FRH]{
  \includegraphics[width=0.3\linewidth]{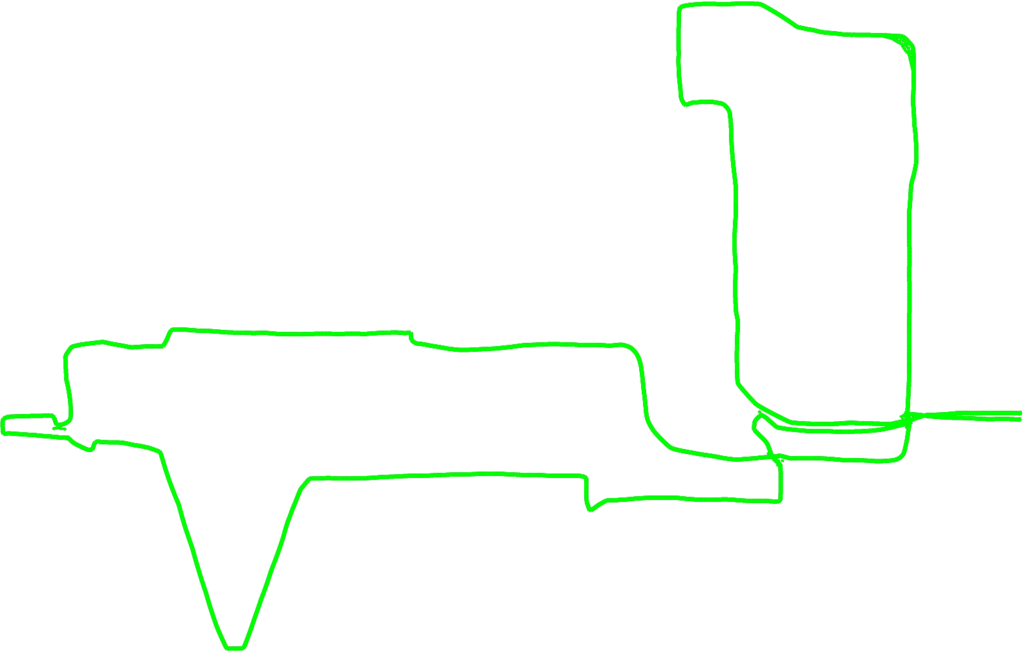} }     
\caption{
Trajectory estimates computed by  \dcMRFc (black) versus ground truth (green)  
for the real datasets CSAIL, FR079, and FRH. \vspace{-5mm}
}
\label{fig:realisticShadowPlots}
\end{figure}

\subsection{Experiments On Real Datasets}
\label{sec:realExperiments}

\myParagraph{Methodology}
In this section, we consider three real-world standard benchmarking datasets, 
the CSAIL dataset 
(1045 poses and 1172 edges),
the FR079 dataset (989 poses and 1217 edges), 
and the FRH dataset 
(1316 poses and 2820 edges). 
We spoiled those datasets with 20 randomly grouped outliers. 
\finalmodification{We add correlation terms with $\barcij = 1$ for each pair of edges connecting consecutive nodes, e.g., $(i,j)$ and $(i\pm1,j\pm1)$.}
We benchmarked our approach against \vertigo, \rrr, and \dcs.
\omitted{
For our technique, we computed the maximum , as described in the appendix \ref{sec:cThreshold}, using the median standard deviation on rotations and translations in the corresponding \scenario{g2o} files. For this experiment, we used the decoupled formulation of our technique (\dcMRFd). For large sized problems, our formulation can be solved using Riemannian
staircase. The implementation details are not included here due to the scope
of this paper. We refer readers to \cite{Rosen17irosws-SEsyncImplementation}
for example implementations of this algorithm. This implementation on Matlab
is still in a preliminary stage, so the execution time (a few hours) is still far
from real time. 
}{}

\begin{table*}
  \centering
  \caption{Average translation error (meters) on real benchmarking datasets} 
  \begin{tabular}{|l||c|c|c|c|c|c|c|c|c|}
    \hline
    &\dcMRFc & \vertigo &\rrr($t_g$=1)&\rrr($t_g$=5)&\rrr($t_g$=10) &\dcs($\Phi=1$) &\dcs($\Phi=10$)&\dcs($\Phi=100$) & \modification{Odometry}
    \begin{filecontents*}{realistic_datasets_table.csv}
Dataset,DC-GM,Vertigo,RRR1,RRR5,RRR10,DCS1,DCS10,DCS100,OdometryN,OdometryY
FRH,0.0008,0.0005,0.0004,0.0003,0.0003,0.0004,0.0004,0.0004,0.0035,0.0021
FR079,0.0438,0.2751,0.0546,0.0520,0.0521,0.2721,0.1804,0.1250,0.5260,0.2836
CSAIL,0.0430,1.4625,0.0495,0.0613,0.0506,1.4576,1.4240,0.0521,2.8730,1.4480
\end{filecontents*}
\csvreader[head to column names]{realistic_datasets_table.csv}{}
    {\\\hline\csvcoli & \modification{\csvcolii} & \modification{\csvcoliii} & \modification{\csvcoliv} & \modification{\csvcolv} & \modification{\csvcolvi} & \modification{\csvcolvii} & \modification{\csvcolviii} & \modification{\csvcolix} & \modification{\csvcolxi}}\\\hline
  \end{tabular}
  \label{tab:realisticGraphsErrors}
  \vspace{-3mm}
\end{table*}

\myParagraph{Results and Interpretation}
Table~\ref{tab:realisticGraphsErrors} presents the average translation error \modification{(computed with respect to the optimized trajectory without outliers)} for all datasets and techniques. \finalmodification{We also report} the average translation error of the odometric estimate.
\modification{All compared techniques achieve very good results on the FRH dataset. This is probably due to the fact that this dataset provides a very good initial guess, hence the techniques that rely on iterative optimization are favored. This intuition is confirmed by the high accuracy of the odometry. The results on the FR079 dataset are more interesting. In this case, \dcMRFc and \rrr achieve the best results with a slight advantage towards \dcMRFc. However, \vertigo performs poorly and \dcs performance remains worse than the proposed technique even with its best parameter choice. \dcMRFc has also the best performance on the CSAIL dataset. Again, \rrr achieves very good results while \vertigo and \dcs have poor performance except for some parameter choice (e.g., \dcs performs well for $\Phi=\{100\}$).}
 We attribute this performance boost to the fact that the proposed approach provides a more direct control on the maximum admissible 
 error \modification{of each measurement}, while the parameters in \dcs and \vertigo have a less clear physical interpretation. 
 This translates to the fact that it is more difficult for \dcs and \vertigo to strike a balance between outlier rejection and inlier selection. 
 Therefore, even when these approaches are able to discard most outliers, they may lose accuracy since they also tend to discard good measurements.
 \modification{The difficulty in performing parameter tuning for \dcs is confirmed by the fact that the value $\Phi=1$
 (recommended by Agarwal~\setal\cite{Agarwal13icra}) 
 leads to   
good results on FRH, but fails on FR079 and CSAIL.} 

Fig.~\ref{fig:realisticShadowPlots} shows the trajectory estimates produced by \dcMRFc for the three real datasets, CSAIL, FR079, and FRH. 


\section{Conclusion}
\label{sec:conclusion}

We introduced a discrete-continuous graphical model (\DCGM) to capture perceptual aliasing and outlier correlation in 
SLAM. 
 Then we developed a semidefinite (SDP) relaxation to perform near-optimal inference in the \DCGM 
 and obtain robust SLAM estimates. 
  Our experiments show that the proposed approach compares favorably with the 
 state of the art  \modification{while not relying on an initial guess for optimization. Our approach also enables a more intuitive tuning of the parameters (e.g., the maximum admissible residual $\barc$). The supplemental material~\cite{Lajoie18tr-DCGM} contains extra results to provide
 more insights on the performance and limitations of the proposed approach.
 }
  %
 This paper opens several avenues for future work. 
 First, our Matlab implementation is currently slow: we plan to develop specialized solvers to optimize the SDP relaxations presented in this paper efficiently, 
 leveraging previous work~\cite{Rosen16wafr-sesync}.
 Second, we plan to extend our testing to 3D SLAM problems: 
 the mathematical formulation in this paper is general, while for numerical reasons we had to limit our tests to relatively small 2D problems.
 \modification{Third, it would be useful to develop incremental solvers that can re-use computation when the measurements are presented to the robot in online (rather than batch) fashion.}
  Finally, it would be interesting to provide a theoretical bound on the number of outliers the proposed technique can tolerate.
\appendix

This appendix proves Proposition~\ref{prop:matrix} by showing
how to reformulate problem~\eqref{eq:robustPGOsplit} using the matrix $\MX$ in~\eqref{eq:X}. 
Let us start by rewriting problem \eqref{eq:robustPGOsplit} \modification{and replacing} the (scalar) discrete variables $\bmp_{ij} \in \{-1,+1\}$ 
with ``binary'' selection matrices $\Bmp_{ij} \in \{-\MI_d; +\MI_d\}$: 
\begin{equation}
\label{eq:robustPGOmatrix}
\begin{split}
& \!\!\displaystyle\min_{ \substack{  \MT_i \in \SOd \times \Real{d}  \\ \Bmp_{ij} \in \{-\MI_d;+\MI_d\} } }  
\!\!\displaystyle \!\!\!\! \sum_{\;\;\;\;(i,j) \in  \calE_{od}}  \| \resPose \|_\MOmega^2 \\
& \finalmodification{-\;\; \!\!\!\!\displaystyle \!\!\!\! \sum_{(i,j) \in \calE_{lc}} \frac{\barc}{2d}\trace{\Bmp_{ij}}
- \!\!\!\! \displaystyle\!\!\!\! \sum_{\;\;\;\;(i,j), (i',j') \in \calP}  
\!\!\!\!\!\!\frac{\barcij}{d} \trace{\Bmp_{ij}\tran\Bmp_{i'j'}}} \\
& + \!\!\!\!\displaystyle \!\!\!\! \sum_{\;\;(i,j) \in \calE_{lc}\;\;} 
\frac{1}{4} \| \resPose + \Bmp_{ij}\tran(\resPose) \|_\MOmega^2  
\end{split}
\end{equation}
where we also rearranged the summands.
Note the division by $d$ in the second and third sum  in~\eqref{eq:robustPGOmatrix}, needed to 
compensate for the fact that we are now working with $d \times d$ matrices \modification{$\Bmp_{ij}$}. 

\finalmodification{The first summation} in~\eqref{eq:robustPGOmatrix} can be written as
\begin{equation}
\label{eq:linear1}
\begin{split}
& \displaystyle \!\!\!\! \sum_{\;\;\;\;(i,j) \in \calE_{od}}  \| \resPose \|_\MOmega^2 \\
& \modification{=\displaystyle\!\!\!\! \sum_{\;\;\;\;(i,j) \in \calE_{od}} \trace{(\resPose) \MOmega (\resPose)\tran}}\\
& = \trace{\M{L}(\calG_{od})\MT\tran\MT}
\end{split}
\end{equation}
where $\M{L}(\calG_{od}) \in \Real{(d+1)n \times (d+1)n}$ is the \textit{Connection Laplacian}~\cite{Rosen16wafr-sesync} of the graph $\calG_{od} = (\calV, \calE_{od})$, 
which has the same set of nodes $\calV$ as the original pose graph, but only includes odometric edges $\calE_{od}$. 
 We can use a derivation similar to~\cite{Briales17ral} to show that the \textit{Connection Laplacian}  
 of a generic graph $\calG = (V,E)$ can be written as 
 \beq
 \M{L}(\calG)=\MA(\veccalG)\MOmega(\calG)\MA(\veccalG)\tran
 \eeq 
  where, 
  the matrices $\MA(\veccalG) \in \Real{(d+1)|V| \times (d+1)|E|}$ 
  and $\MOmega(\calG) \in \Real{(d+1)|E| \times (d+1)|E|}$ 
  are given as follows:
\begin{equation}
\finalmodification{
\;[\MA(\veccalG)]_{r,e} \doteq \left\{
\begin{array}{ll}
\!\!\!\!-{\bar{\MT}}_{i_e, j_e} &\!\!\!\!\text{if } r=i_e,\\
\!\!\!\!+\MI_{d+1} &\!\!\!\!\text{if } r = j_e,\\
\!\!\!\!\M{0}_{d+1} &\!\!\!\! \text{otherwise.}
\end{array}\right.  \text{ for } e=1,\ldots,|E| }
\end{equation}
\begin{equation}
\hspace{-1cm}\MOmega(\calG) \doteq \text{blkdiag}(\MOmega_1, ... , \MOmega_{|E|})
\end{equation}
The notation $[\MA(\veccalG)]_{r,e}$ denotes the $(d+1)\times(d+1)$ block of $\MA(\veccalG)$ 
at block row $r$ and block column $e$, while the $e$-th edge in $E$ is denoted as $(i_e,j_e)$. 

The second summation in~\eqref{eq:robustPGOmatrix} can be developed as follows:
\beq
\label{eq:linear2}
\displaystyle \sum_{\;\;\;\;(i,j) \in \calE_{lc}} 
- \frac{\barc}{2d}\trace{\Bmp_{ij}} = - \frac{\barc}{2d}\trace{\MI_{d, \nrLoops}\tran\Bmp} 
\eeq
where $\modification{\MI_{d, \nrLoops}}$ 
$\doteq 
[\MI_d \, \ldots \, \MI_d]$ is a row of $\nrLoops$ identity matrices. 

Similarly, the third summation in~\eqref{eq:robustPGOmatrix} can be written as:
\begin{equation}
\label{eq:linear3}
\finalmodification{\displaystyle -\hspace{-5mm}\sum_{\;\;\;\;(i,j), (i',j') \in \calP}  
\!\!\!\!\!\!\!\!\!\!\!\!\frac{\barcij}{d} \trace{\Bmp_{ij}\tran\Bmp_{i'j'}} = 
-\frac{1}{2d} \trace{ \MN(\corSet) \Bmp\tran\Bmp}}
\end{equation}
where $\MN(\corSet) \in \Real{ d\nrLoops\times d\nrLoops}$ has $d\times d$ blocks in the form:
\begin{equation}\finalmodification{
\!\!\![\MN(\corSet)]_{e,e'}
\doteq \left\{\begin{array}{ll}
\!\!\barc_{(i'j')}^{(ij)}\MI_d & \!\!\text{if } e = (i,j), e'=(i',j')\in \corSet\\
\!\!\barc_{(ij)}^{(i'j')}\MI_d & \!\!\text{if } e = (i',j'), e'=(i,j)\in \corSet
\\
\!\!\M{0}_d & \!\!\text{otherwise.}
\end{array}\right.}
\end{equation}

The first three terms in~\eqref{eq:robustPGOmatrix} are linear with respect to parts of the matrix $\MX\tran\MX$ in~\eqref{eq:X}, 
so we write the sum of \eqref{eq:linear1},~\eqref{eq:linear2},~\eqref{eq:linear3} compactly as $\trace{\ldata\MX\tran\MX}$ where 
\begin{equation}
\ldata = \begin{bmatrix}
\M{L}(\calG_{od}) & \M{0}_{(d+1)n, d\nrLoops} & \M{0}_{(d+1)n, d} \\
\M{0}_{d\nrLoops, (d+1)n} & \modification{-\frac{1}{2d}\MN(\corSet)} & -\frac{\barc}{4d}\MI_{d,\nrLoops}\tran \\
\M{0}_{d, (d+1)n} & -\frac{\barc}{4d}\MI_{d,\nrLoops} & \M{0}_{d, d}
\end{bmatrix}
\end{equation}
which is the first term in eq.~\eqref{eq:matrixProblem}. Here $\M{0}_{p, q}$ denotes a zero matrix of size $p\times q$.

In order to complete the proof, we only need to show that the last sum in~\eqref{eq:robustPGOmatrix} 
can be written as $\sum_{e=(i,j) \in \calE_{lc}} \!\!\!\!
\trace{\qdataOne\MX\tran\MX\qdataTwo\MX\tran\MX}$, \cf~\eqref{eq:matrixProblem}.  
Towards this goal, we
 develop each squared norm in the last sum 
using a derivation similar to~\eqref{eq:linear1} and get:
\begin{equation}
\label{eq:lc_cost}
\begin{split}
&\;\;\;\;\| \resPose + \Bmp_{ij}\tran(\resPose) \|_\MOmega^2 \\
& = \trace{\M{L}(\calG_{e})\MT\tran\MT} + \trace{\M{L}(\calG_{e})(\Bmp_{ij}\tran\MT)\tran(\Bmp_{ij}\tran\MT)} \\
& + \trace{\M{L}(\calG_{e})(\Bmp_{ij}\tran\MT)\tran\MT} + \trace{\M{L}(\calG_{e})\MT\tran(\Bmp_{ij}\tran\MT)}
\end{split}
\end{equation}
where $\calG_e = (\calV, \calE_{ij})$ denotes a graph with a single edge $e=(i, j)$. 
We can write $\MT$ and $\Bmp\tran\MT$ as matrix blocks in $\MX\tran\MX$: 
\begin{equation}
\begin{split}
\MT =& [\M{0}_{d, (d+1)n+d\nrLoops}\;\;\MI_d]\MX\tran\MX\\
&[\MI_{(d+1)n}\;\;\M{0}_{((d+1)n, d(\nrLoops+1))}]\tran\\
\Bmp_{ij}\tran\MT =& [\M{0}_{d, (d+1)n+d(e-1)}\;\;\MI_d\;\;\M{0}_{d, d(\nrLoops-e+1)}]
\MX\tran\MX\\
&[\MI_{(d+1)n}\;\;\M{0}_{((d+1)n, d(\nrLoops+1))}]\tran
\end{split}
\end{equation}
which enables to write each 
squared norm in terms of $\MX\tran\MX$ as follows:
\begin{equation}
\label{eq:quadratic}
\| \resPose + \Bmp_{ij}\tran(\resPose) \|_\MOmega^2 = \trace{\qdataOne\MX\tran\MX\qdataTwo\MX\tran\MX}
\end{equation}
where

\vspace{-0.5cm}

\begin{equation}
\begin{split}
\qdataOne = &\begin{bmatrix}
\M{L}(\calG_{e}) & \M{0}_{(d+1)n, d\nrLoops} & \M{0}_{(d+1)n, d} \\
\M{0}_{d\nrLoops, (d+1)n} & \M{0}_{d\nrLoops, d\nrLoops} & \M{0}_{d\nrLoops, d} \\
\M{0}_{d, (d+1)n} & \M{0}_{d, d\nrLoops} & \M{0}_{d, d}
\end{bmatrix} \\
\qdataTwo = &[\M{0}_{d, (d+1)n+d(e-1)}\;\;\MI_d\;\;\M{0}_{d, d(m-e)}\;\;\MI_d]\tran \\
& [\M{0}_{d, (d+1)n+d(e-1)}\;\;\MI_d\;\;\M{0}_{d, d(m-e)}\;\;\MI_d]
\end{split}
\end{equation}
Summing over all loop-closure edges results in the second term in eq.~\eqref{eq:matrixProblem}, concluding the proof.


\bibliographystyle{IEEEtran}
\bibliography{../../references/refs,../../references/myRefs}

\begin{thebibliography}{10}
\providecommand{\url}[1]{#1}
\csname url@samestyle\endcsname
\providecommand{\newblock}{\relax}
\providecommand{\bibinfo}[2]{#2}
\providecommand{\BIBentrySTDinterwordspacing}{\spaceskip=0pt\relax}
\providecommand{\BIBentryALTinterwordstretchfactor}{4}
\providecommand{\BIBentryALTinterwordspacing}{\spaceskip=\fontdimen2\font plus
\BIBentryALTinterwordstretchfactor\fontdimen3\font minus
  \fontdimen4\font\relax}
\providecommand{\BIBforeignlanguage}[2]{{%
\expandafter\ifx\csname l@#1\endcsname\relax
\typeout{** WARNING: IEEEtran.bst: No hyphenation pattern has been}%
\typeout{** loaded for the language `#1'. Using the pattern for}%
\typeout{** the default language instead.}%
\else
\language=\csname l@#1\endcsname
\fi
#2}}
\providecommand{\BIBdecl}{\relax}
\BIBdecl

\bibitem{Blake11book-MRF}
A.~Blake, P.~Kohli, and C.~Rother, \emph{{Markov Random Fields} for Vision and
  Image Processing}.\hskip 1em plus 0.5em minus 0.4em\relax The MIT Press,
  2011.

\bibitem{Carlone15icra-verification}
L.~Carlone and F.~Dellaert, ``Duality-based verification techniques for {2D
  SLAM},'' in \emph{IEEE Intl. Conf. on Robotics and Automation (ICRA)}, 2015,
  pp. 4589--4596,
  \linkToPdf{https://www.dropbox.com/s/4wtxyp817hdfnna/2015c-ICRA-duality2D.pdf?dl=0}
  \linkToCode{https://www.bitbucket.org/lucacarlone/pgo2d-duality-opencode}.

\bibitem{Carlone16tro-duality2D}
L.~Carlone, G.~Calafiore, C.~Tommolillo, and F.~Dellaert, ``Planar pose graph
  optimization: Duality, optimal solutions, and verification,'' \emph{{IEEE}
  Trans. Robotics}, vol.~32, no.~3, pp. 545--565, 2016,
  \linkToPdf{https://www.dropbox.com/s/peoktkct0cw42av/2015j-TRO-dualityPGO2D.pdf?dl=0}
  \linkToCode{https://www.bitbucket.org/lucacarlone/pgo2d-duality-opencode}.

\bibitem{Carlone15iros-duality3D}
L.~Carlone, D.~Rosen, G.~Calafiore, J.~Leonard, and F.~Dellaert, ``Lagrangian
  duality in {3D SLAM}: Verification techniques and optimal solutions,'' in
  \emph{IEEE/RSJ Intl. Conf. on Intelligent Robots and Systems (IROS)}, 2015,
  pp. 125--132,
  \linkToPdf{https://www.dropbox.com/s/n4qvlz70ajjfqeb/2015c-iros-duality3D.pdf?dl=0}
  \linkToCode{https://www.bitbucket.org/lucacarlone/pgo3d-duality-opencode}
  (datasets: \linkToWeb{https://lucacarlone.mit.edu/datasets/}) (supplemental
  material:
  \linkToPdf{https://www.dropbox.com/s/bztm41vbo5249i2/2015c-iros-duality3D-supplemental.pdf?dl=0}).

\bibitem{Carlone18ral-robustPGO2D}
L.~Carlone and G.~Calafiore, ``Convex relaxations for pose graph optimization
  with outliers,'' \emph{{IEEE} Robotics and Automation Letters ({RA-L})},
  vol.~3, no.~2, pp. 1160--1167, 2018, arxiv preprint: 1801.02112,
  \linkToPdf{https://arxiv.org/pdf/1801.02112.pdf}.

\bibitem{Rosen16wafr-sesync}
D.~Rosen, L.~Carlone, A.~Bandeira, and J.~Leonard, ``{SE-Sync}: A certifiably
  correct algorithm for synchronization over the {Special Euclidean} group,''
  in \emph{Intl. Workshop on the Algorithmic Foundations of Robotics (WAFR)},
  San Francisco, CA, December 2016, extended arxiv preprint: 1611.00128,
  \linkToPdf{http://arxiv.org/abs/1611.00128}
  \linkToPdf{http://wafr2016.berkeley.edu/papers/WAFR_2016_paper_138.pdf}
  \linkToCode{https://github.com/david-m-rosen/SE-Sync}\award{, best paper
  award}.

\bibitem{Sunderhauf12iros}
N.~S\"{u}nderhauf and P.~Protzel, ``Switchable constraints for robust pose
  graph {SLAM},'' in \emph{IEEE/RSJ Intl. Conf. on Intelligent Robots and
  Systems (IROS)}, 2012.

\bibitem{Latif12rss}
Y.~Latif, C.~D.~C. Lerma, and J.~Neira, ``Robust loop closing over time.'' in
  \emph{Robotics: Science and Systems (RSS)}, 2012.

\bibitem{Agarwal13icra}
P.~Agarwal, G.~Tipaldi, L.~Spinello, C.~Stachniss, and W.~Burgard, ``Robust map
  optimization using dynamic covariance scaling,'' in \emph{IEEE Intl. Conf. on
  Robotics and Automation (ICRA)}, 2013.

\bibitem{Szeliski08pami-surveyMRF}
R.~Szeliski, R.~Zabih, D.~Scharstein, O.~Veksler, V.~Kolmogorov, A.~Agarwala,
  M.~Tappen, and C.~Rother, ``{A Comparative Study of Energy Minimization
  Methods for Markov Random Fields with Smoothness-Based Priors},'' \emph{IEEE
  Transactions on Pattern Analysis and Machine Intelligence}, vol.~30, no.~6,
  pp. 1068--1080, 2008.

\bibitem{Kappes15ijcv-energyMin}
J.~H. Kappes, B.~Andres, F.~A. Hamprecht, C.~Schn{\"o}rr, S.~Nowozin, D.~Batra,
  S.~Kim, B.~X. Kausler, T.~Kr{\"o}ger, J.~Lellmann, N.~Komodakis,
  B.~Savchynskyy, and C.~Rother, ``{A Comparative Study of Modern Inference
  Techniques for Structured Discrete Energy Minimization Problems},''
  \emph{Intl. J. of Computer Vision}, vol. 115, no.~2, pp. 155--184, 2015.

\bibitem{Fix14eccv}
A.~Fix and S.~Agarwal, ``Duality and the continuous graphical model,'' in
  \emph{European Conf. on Computer Vision (ECCV)}, 2014, pp. 266--281.

\bibitem{Zach12eccv}
C.~Zach and P.~Kohli, ``A convex discrete-continuous approach for {M}arkov
  random fields,'' in \emph{European Conf. on Computer Vision (ECCV)}, 2012,
  pp. 386--399.

\bibitem{Crandall12pami}
D.~Crandall, A.~Owens, N.~Snavely, and D.~Huttenlocher, ``{SfM} with {MRFs}:
  Discrete-continuous optimization for large-scale structure from motion,''
  \emph{{IEEE} Trans. Pattern Anal. Machine Intell.}, 2012.

\bibitem{Hartley13ijcv}
R.~Hartley, J.~Trumpf, Y.~Dai, and H.~Li, ``Rotation averaging,'' \emph{IJCV},
  vol. 103, no.~3, pp. 267--305, 2013.

\bibitem{Dellaert05rss}
F.~Dellaert, ``Square {Root} {SAM}: Simultaneous location and mapping via
  square root information smoothing,'' in \emph{Robotics: Science and Systems
  (RSS)}, 2005.

\bibitem{Huber81}
P.~Huber, \emph{Robust Statistics}.\hskip 1em plus 0.5em minus 0.4em\relax John
  Wiley \& Sons, New York, NY, 1981.

\bibitem{Bosse17fnt}
M.~Bosse, G.~Agamennoni, and I.~Gilitschenski, ``Robust estimation and
  applications in robotics,'' \emph{Foundations and Trends in Robotics},
  vol.~4, no.~4, pp. 225--269, 2016.

\bibitem{Olson12rss}
E.~Olson and P.~Agarwal, ``Inference on networks of mixtures for robust robot
  mapping,'' in \emph{Robotics: Science and Systems (RSS)}, July 2012.

\bibitem{Casafranca13iros}
J.~Casafranca, L.~Paz, and P.~Pini{\'e}s, ``A back-end $\ell_1$ norm based
  solution for factor graph {SLAM},'' in \emph{IEEE/RSJ Intl. Conf. on
  Intelligent Robots and Systems (IROS)}, 2013, pp. 17--23.

\bibitem{Lee13iros}
G.~H. Lee, F.~Fraundorfer, and M.~Pollefeys, ``Robust pose-graph loop-closures
  with expectation-maximization,'' in \emph{IEEE/RSJ Intl. Conf. on Intelligent
  Robots and Systems (IROS)}, 2013.

\bibitem{Fischler81}
M.~Fischler and R.~Bolles, ``Random sample consensus: a paradigm for model
  fitting with application to image analysis and automated cartography,''
  \emph{Commun. ACM}, vol.~24, pp. 381--395, 1981.

\bibitem{Neira01tra}
J.~Neira and J.~Tard{\'o}s, ``Data association in stochastic mapping using the
  joint compatibility test,'' \emph{{IEEE} Trans. Robot. Automat.}, vol.~17,
  no.~6, pp. 890--897, December 2001.

\bibitem{Sunderhauf12icra}
N.~Sunderhauf and P.~Protzel, ``Towards a robust back-end for pose graph
  {SLAM},'' in \emph{IEEE Intl. Conf. on Robotics and Automation (ICRA)}, 2012,
  pp. 1254--1261.

\bibitem{Carlone14iros-robustPGO2D}
L.~Carlone, A.~Censi, and F.~Dellaert, ``Selecting good measurements via
  $\ell_1$ relaxation: a convex approach for robust estimation over graphs,''
  in \emph{IEEE/RSJ Intl. Conf. on Intelligent Robots and Systems (IROS)},
  2014,
  \linkToPdf{https://www.dropbox.com/s/7f304d5ag245ie4/2014c-IROS-outlierRejection.pdf?dl=0}.

\bibitem{Graham15iros}
M.~Graham, J.~How, and D.~Gustafson, ``Robust incremental {SLAM} with
  consistency-checking,'' in \emph{IEEE/RSJ Intl. Conf. on Intelligent Robots
  and Systems (IROS)}, Sept 2015, pp. 117--124.

\bibitem{Mangelson18icra}
J.~Mangelson, D.~Dominic, R.~Eustice, and R.~Vasudevan, ``Pairwise consistent
  measurement set maximization for robust multi-robot map merging,'' in
  \emph{IEEE Intl. Conf. on Robotics and Automation (ICRA)}, 2018.

\bibitem{Pfingsthorn13ijrr}
M.~Pfingsthorn and A.~Birk, ``Simultaneous localization and mapping with
  multimodal probability distributions,'' \emph{Intl. J. of Robotics Research},
  vol.~32, no.~2, pp. 143--171, 2013.

\bibitem{Pfingsthorn16ijrr}
------, ``Generalized graph {SLAM}: Solving local and global ambiguities
  through multimodal and hyperedge constraints,'' \emph{Intl. J. of Robotics
  Research}, vol.~35, no.~6, pp. 601--630, 2016.

\bibitem{Bowman17icra}
S.~Bowman, N.~Atanasov, K.~Daniilidis, and G.~Pappas, ``Probabilistic data
  association for semantic slam,'' in \emph{IEEE Intl. Conf. on Robotics and
  Automation (ICRA)}, 2017, pp. 1722--1729.

\bibitem{Dellaert00cvpr}
F.~Dellaert, S.~Seitz, C.~Thorpe, and S.~Thrun, ``Structure from motion without
  correspondence,'' in \emph{IEEE Conf. on Computer Vision and Pattern
  Recognition (CVPR)}, June 2000.

\bibitem{Wang13ima}
L.~Wang and A.~Singer, ``Exact and stable recovery of rotations for robust
  synchronization,'' \emph{Information and Inference: A Journal of the IMA},
  vol.~30, 2013.

\bibitem{Arrigoni18cviu}
F.~Arrigoni, B.~Rossi, P.~Fragneto, and A.~Fusiello, ``Robust synchronization
  in {SO(3)} and {SE(3)} via low-rank and sparse matrix decomposition,''
  \emph{Computer Vision and Image Understanding}, 2018.

\bibitem{Briales17ral}
J.~Briales and J.~Gonzalez-Jimenez, ``Cartan-sync: Fast and global
  {SE(d)}-synchronization,'' \emph{IEEE Robot. Autom. Lett}, vol.~2, no.~4, pp.
  2127--2134, 2017.

\bibitem{Lajoie18tr-DCGM}
P.~Lajoie, S.~Hu, G.~Beltrame, and L.~Carlone, ``Modeling perceptual aliasing
  in {SLAM} via discrete-continuous graphical models,'' Tech. Rep., 2018, arXiv
  preprint: \linkToPdf{https://arxiv.org/pdf/1810.11692.pdf}, Supplemental
  Material:
  \linkToPdf{https://www.dropbox.com/s/vupak65wi75yzbl/2018j-RAL-DCGM-supplemental.pdf?dl=0}.

\bibitem{Tron15rssws3D-dualityPGO3D}
R.~Tron, D.~Rosen, and L.~Carlone, ``On the inclusion of determinant
  constraints in lagrangian duality for {3D SLAM},'' in \emph{Robotics: Science
  and Systems (RSS), Workshop ``The problem of mobile sensors: Setting future
  goals and indicators of progress for {SLAM}''}, 2015,
  \linkToPdf{https://www.dropbox.com/s/859umrdf7ldd2kv/2015ws-rss-duality3Ddet.pdf?dl=0}.

\bibitem{CVXwebsite}
\BIBentryALTinterwordspacing
M.~Grant and S.~Boyd, ``{CVX}: Matlab software for disciplined convex
  programming.'' [Online]. Available: \url{http://cvxr.com/cvx}
\BIBentrySTDinterwordspacing

\end{thebibliography}
\newpage
\onecolumn

\bgroup\par\addvspace{0.5\baselineskip}\centering%
\normalfont\normalsize\vskip0.2em{\Huge Modeling Perceptual Aliasing in SLAM \\ via Discrete-Continuous Graphical Models \\\smallskip \Large{-- Supplemental material --  \\\smallskip \normalsize{Pierre-Yves Lajoie, Siyi Hu, Giovanni Beltrame, Luca Carlone}} \par}\vskip1.0em\par
\par\addvspace{0.5\baselineskip}\egroup

This supplemental material presents a set of additional experimental results to provide more insights on the performance and limitations of the proposed approach. 
The experiments are organized in four sections.
The experiments in Section~\ref{sec:barc} evaluate the impact of the maximum admissible residuals parameter $\barc$. 
Section~\ref{sec:barcij} analyzes 
the effect of the correlation terms $\bar{c}^{(ij)}_{(i'j')}$ on the truncated least squares objective function and gives some intuition on how to choose these parameters.
Following this analysis, the experiments in Section~\ref{sec:heterogeneous} show the impact of an incorrect modeling of outlier correlation in \dcMRFc. 
%
Finally, 
Section~\ref{sec:partial_grid} shows 
extra simulation results in a more realistic Manhattan World graph that complement the results on the Grid graph shown in the main paper.
%

\subsection{Effect of the Maximum Admissible Residual Threshold $\barc$}\label{sec:barc}

This set of experiments evaluates the impact  of the choice of the  maximum admissible residuals threshold $\bar{c}$. In order to illustrate the role of $\barc$, we used the same simulation setup of the paper, but we varied the number of standard deviation of the measurement noise ($\sigma$) that we wished to accept. Intuitively, we expect that a lower $\bar{c}$ will lead to more inliers being rejected and a higher $\bar{c}$ will lead to more outliers being accepted. 

We evaluate the results for 3 choices of $\bar{c}$, in particular we consider $\barc = \{0.01\sigma, 1\sigma, 2\sigma\}$, and for each set of tests we report (i) the average translation error, (ii) the percentage of rejected inliers, (iii) the percentage of rejected outliers, and (iv) the rank of the matrix $\MZ^\star$ computed by the proposed SDP relaxation (in planar problems, the relaxation is tight when $\rank{\MZ^\star} = 2$).
 The rank is computed using a numerical threshold of $10^{-3}\cdot\lambda_{max}$, where $\lambda_{max}$ is the maximum eigenvalue of $\MZ^\star$. All data points are averaged over 5 runs, and statistics are computed for increasing percentage of outliers.
\begin{figure}[H]
\centering
\begin{minipage}{0.4\textwidth}
\centering
\subfloat{
	\includegraphics[width=0.8\textwidth]{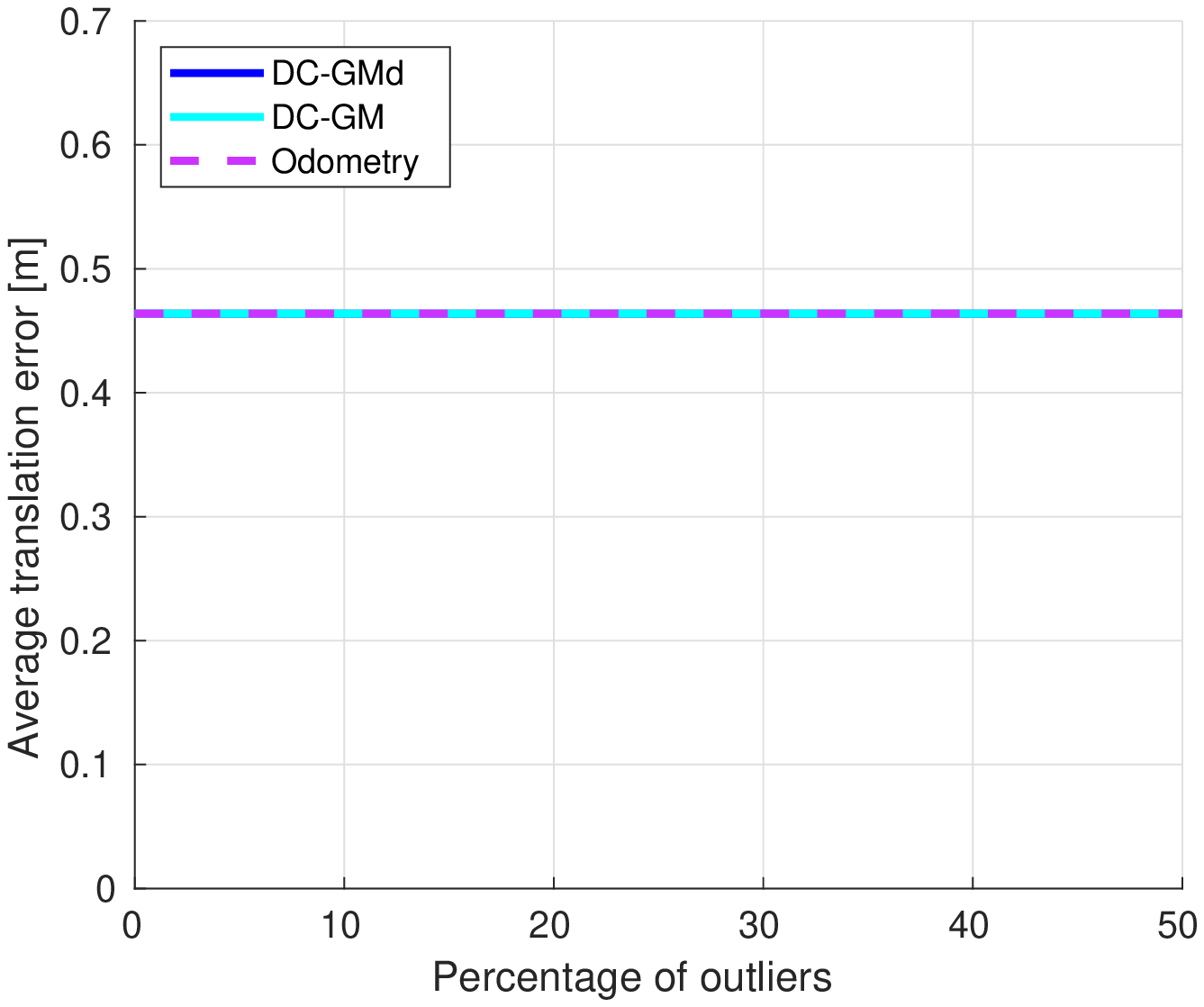}  (a)
}
	
\subfloat{
	\includegraphics[width=0.8\textwidth]{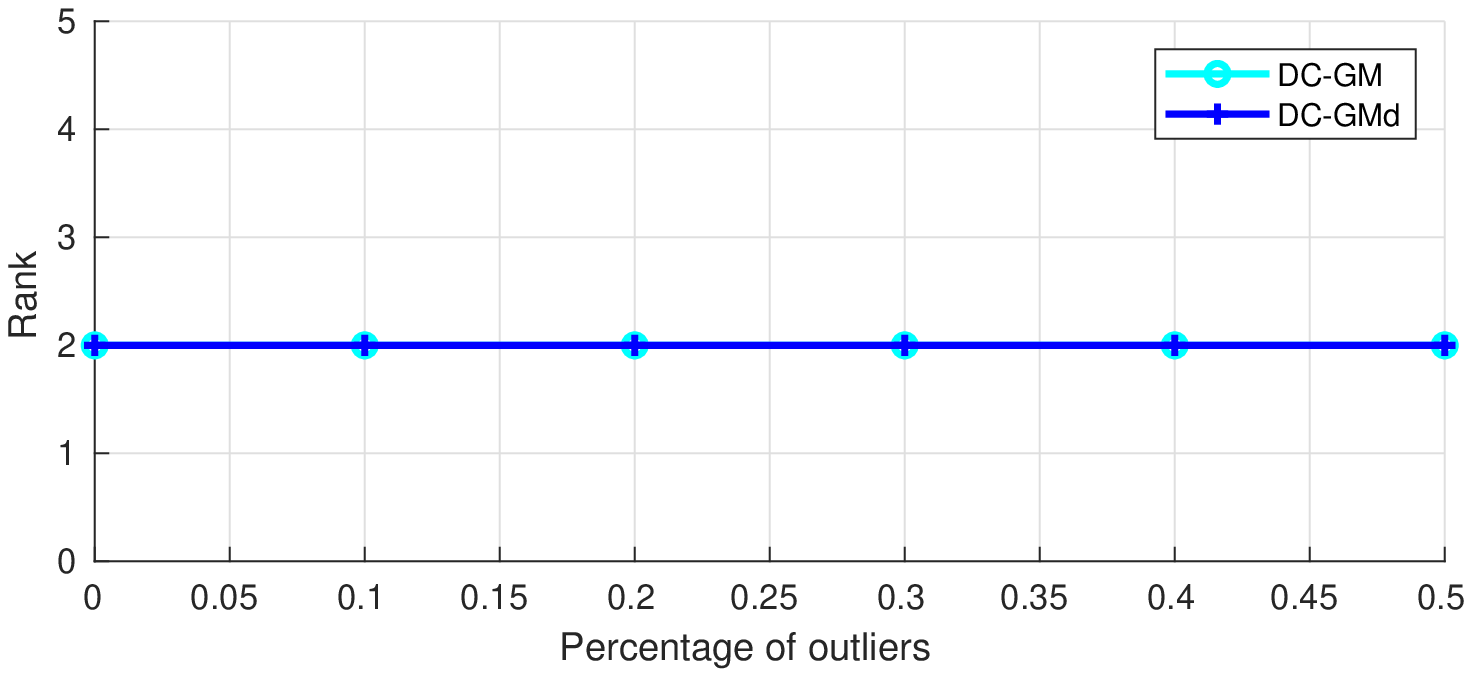}  (b)
}
\end{minipage}
\begin{minipage}{0.4\textwidth}
\centering
\subfloat{
	\includegraphics[width=1\textwidth]{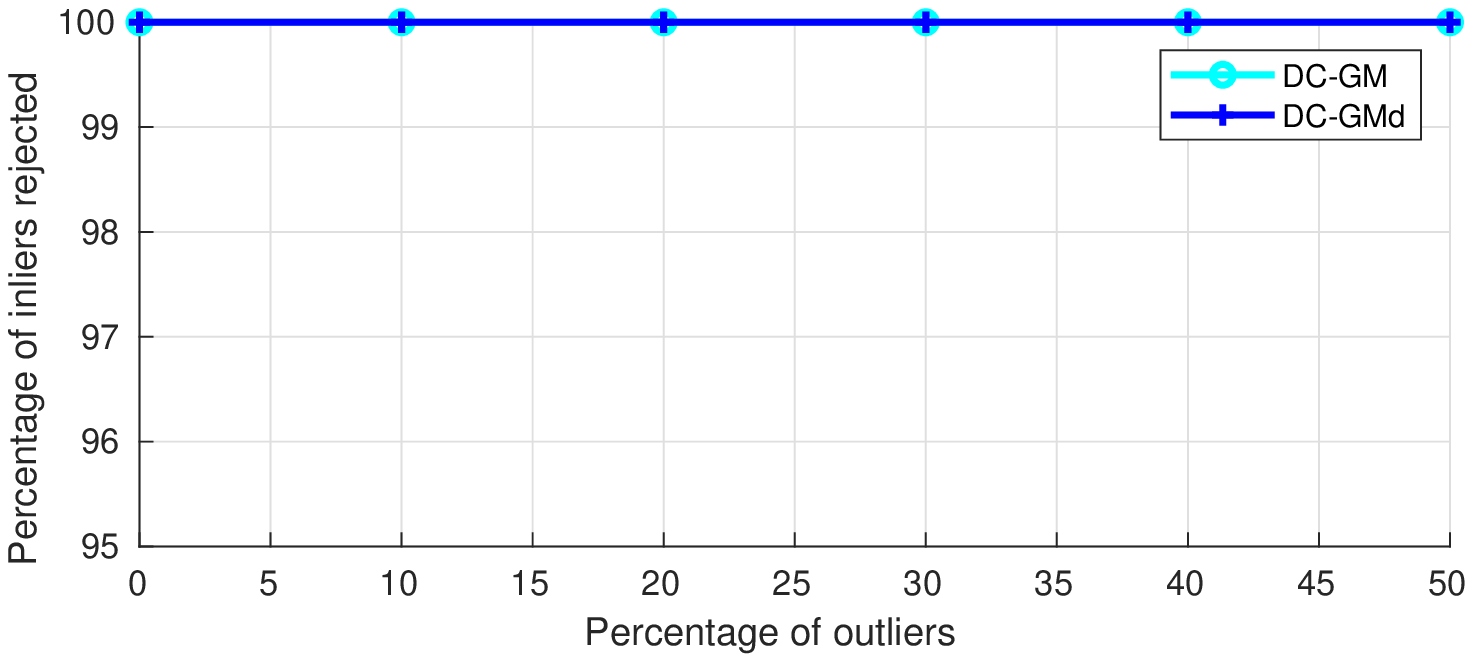} (c)
}
	
\vspace{10mm}
\subfloat{
	\includegraphics[width=1\textwidth]{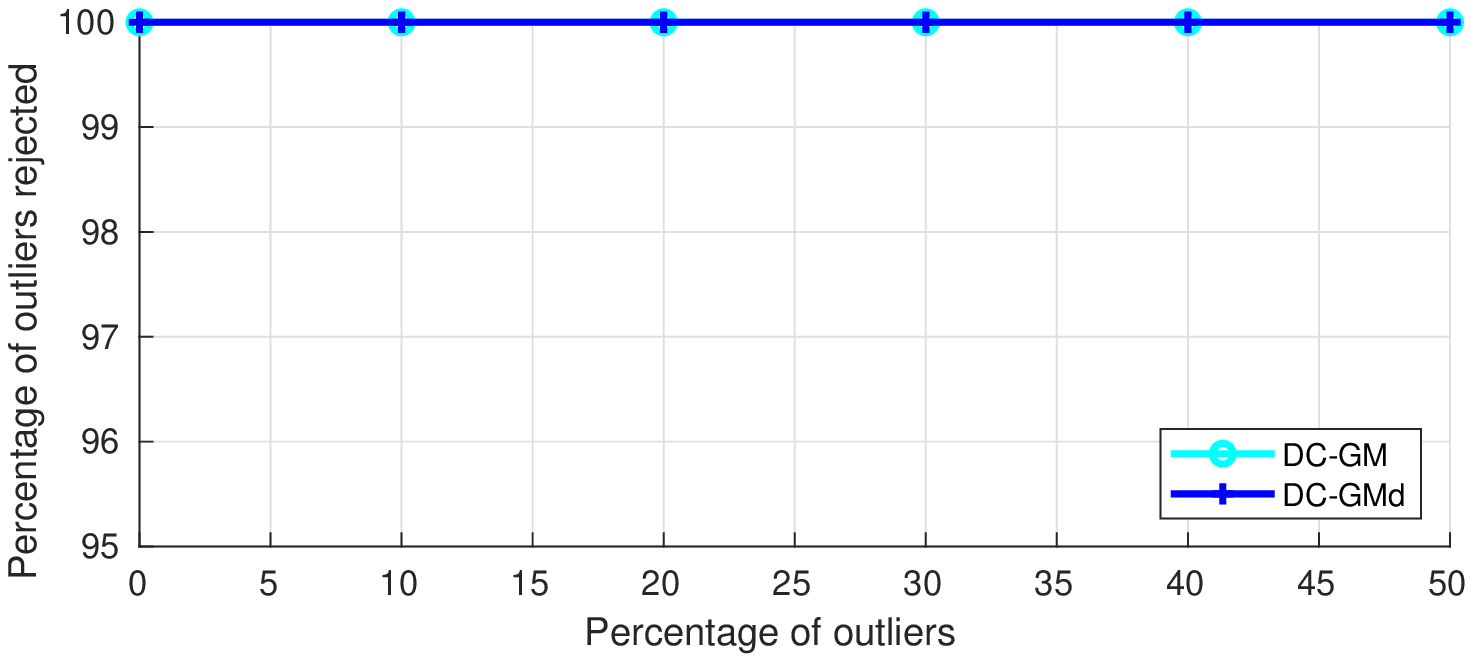} (d)
}
\end{minipage}
\caption{Results on the simulated grid graph with maximum admissible residuals of 0.01$\sigma$. 
(a) average translation error of the \dcMRFc and \dcMRFd solutions compared with the odometric estimate; 
(b) rank of $\MZ^\star$,
(c) percentage of rejected inliers, and
(d) percentage of rejected outliers for \dcMRFc and \dcMRFd.}
\label{fig:nrsigma0_01}
\end{figure}
As expected, in Figure~\ref{fig:nrsigma0_01} we observe that with a very low tolerance on the residuals, our technique rejects all loop closures and therefore falls back to the odometric estimate. It is worth noting that in this case the rank of the returned solution $\MZ^\star$ is exactly 2 which indicates that the relaxation is tight when all loop closures are rejected.
\begin{figure}[H]
\centering
\begin{minipage}{0.4\textwidth}
\centering
\subfloat{
	\includegraphics[width=0.8\textwidth]{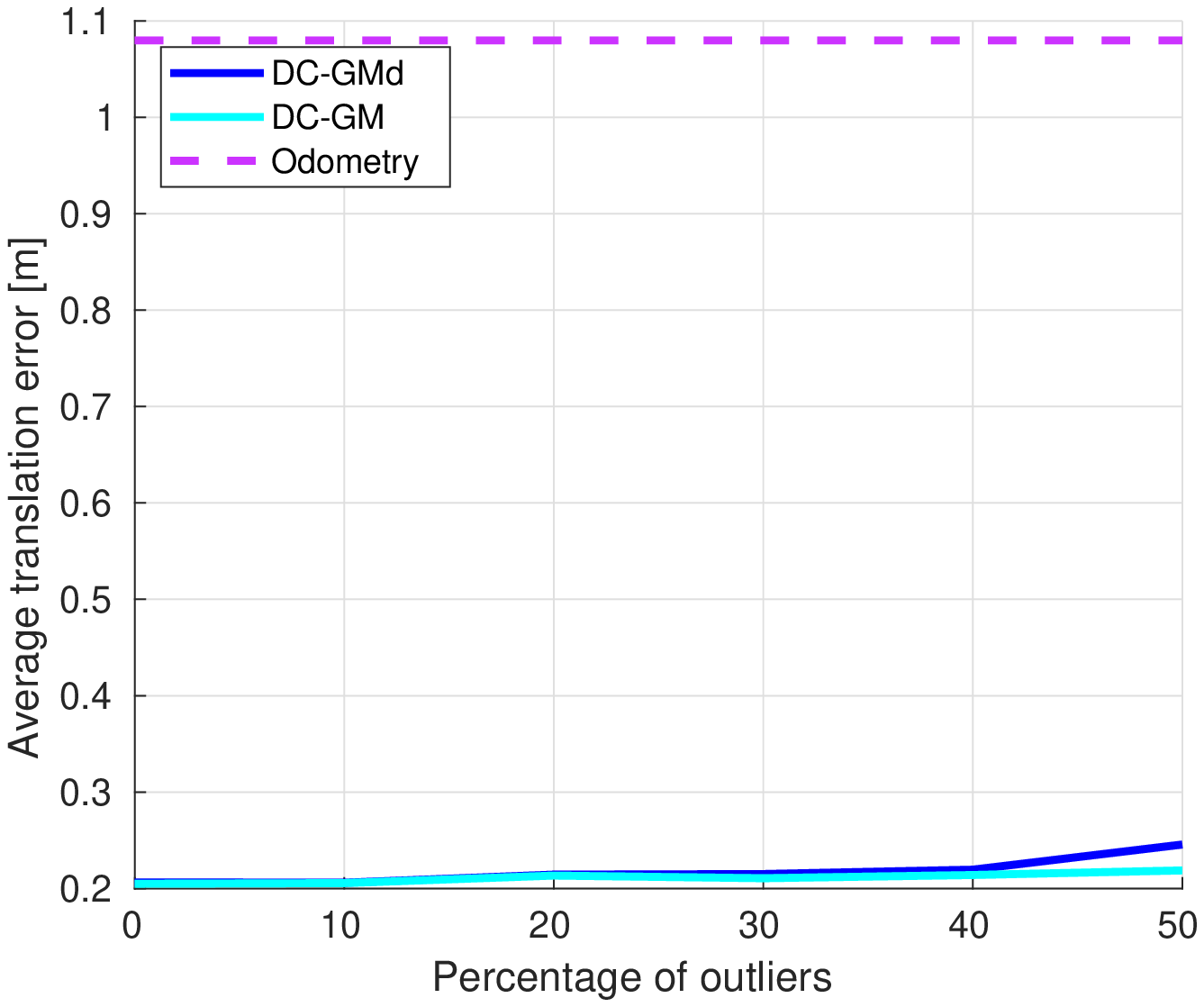} (a)
}
	
\subfloat{
	\includegraphics[width=0.8\textwidth]{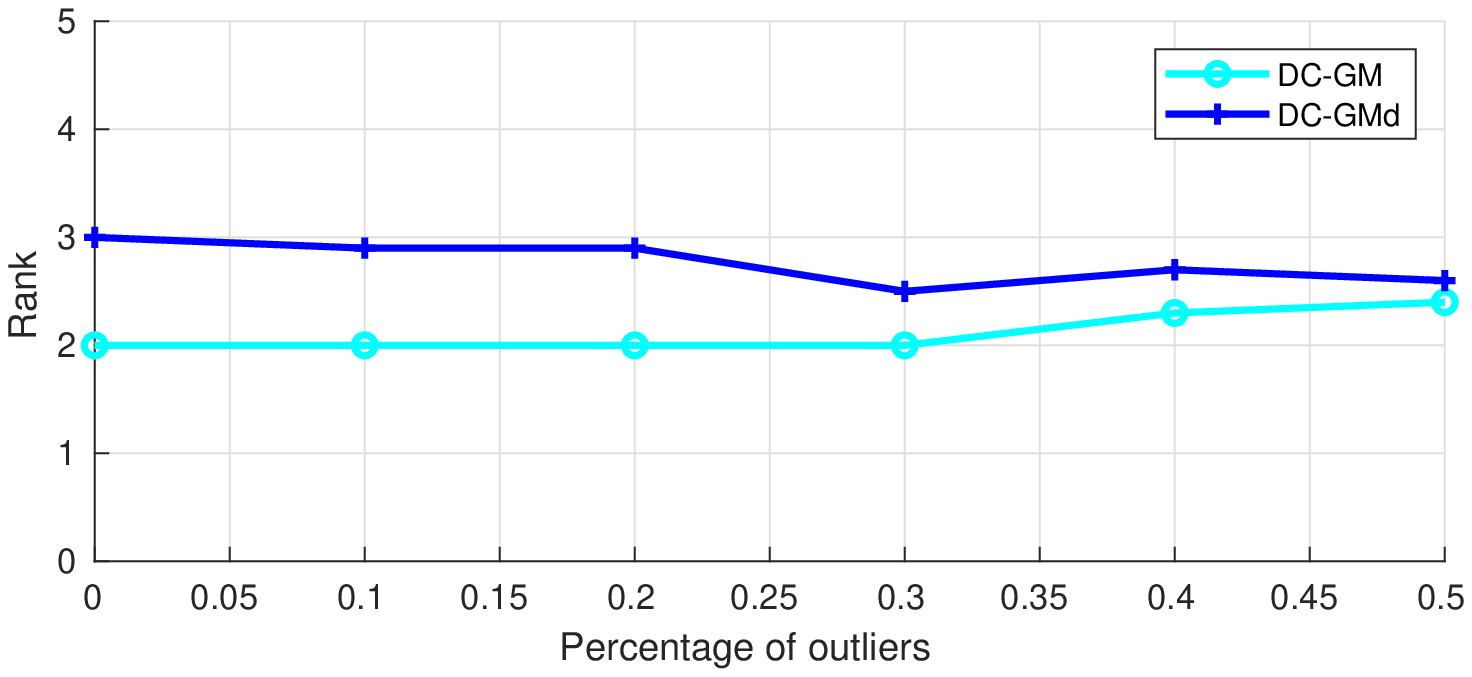} (b)
}
\end{minipage}
\begin{minipage}{0.4\textwidth}
\centering
\subfloat{
	\includegraphics[width=1\textwidth]{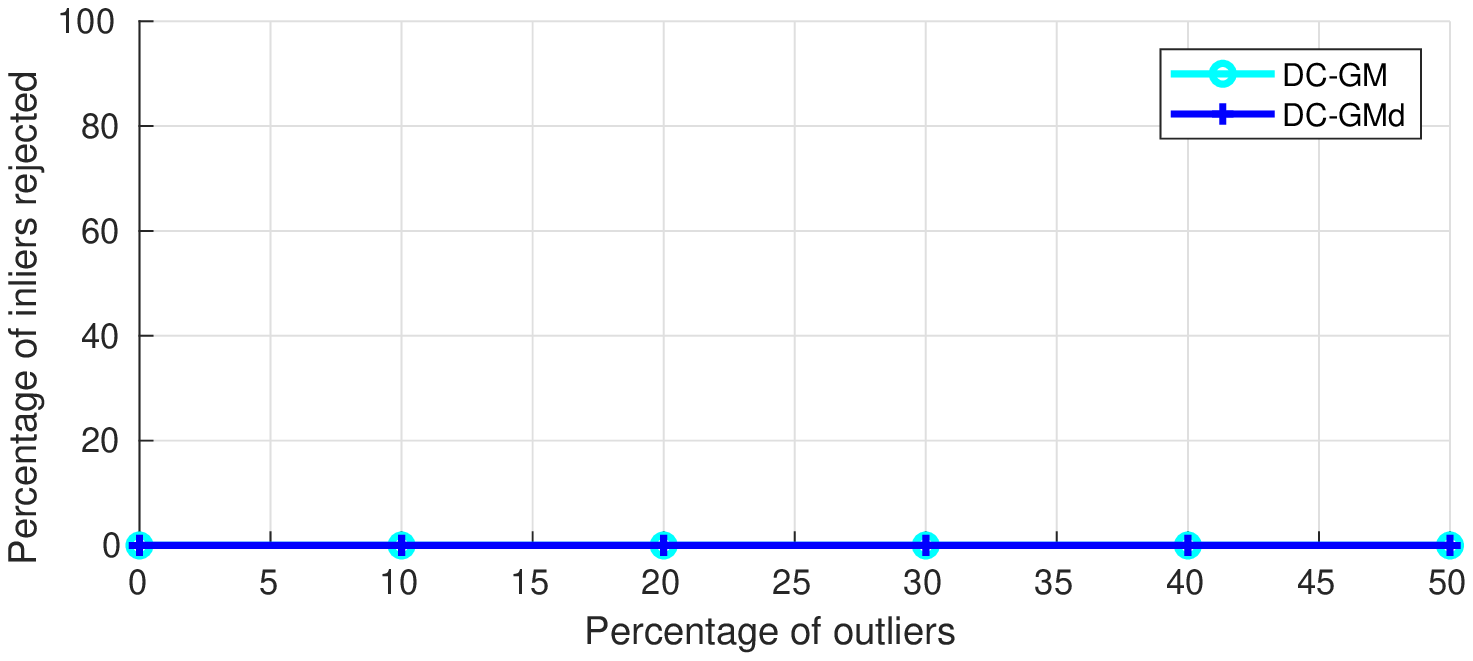} (c)
}
	
\vspace{10mm}
\subfloat{
	\includegraphics[width=1\textwidth]{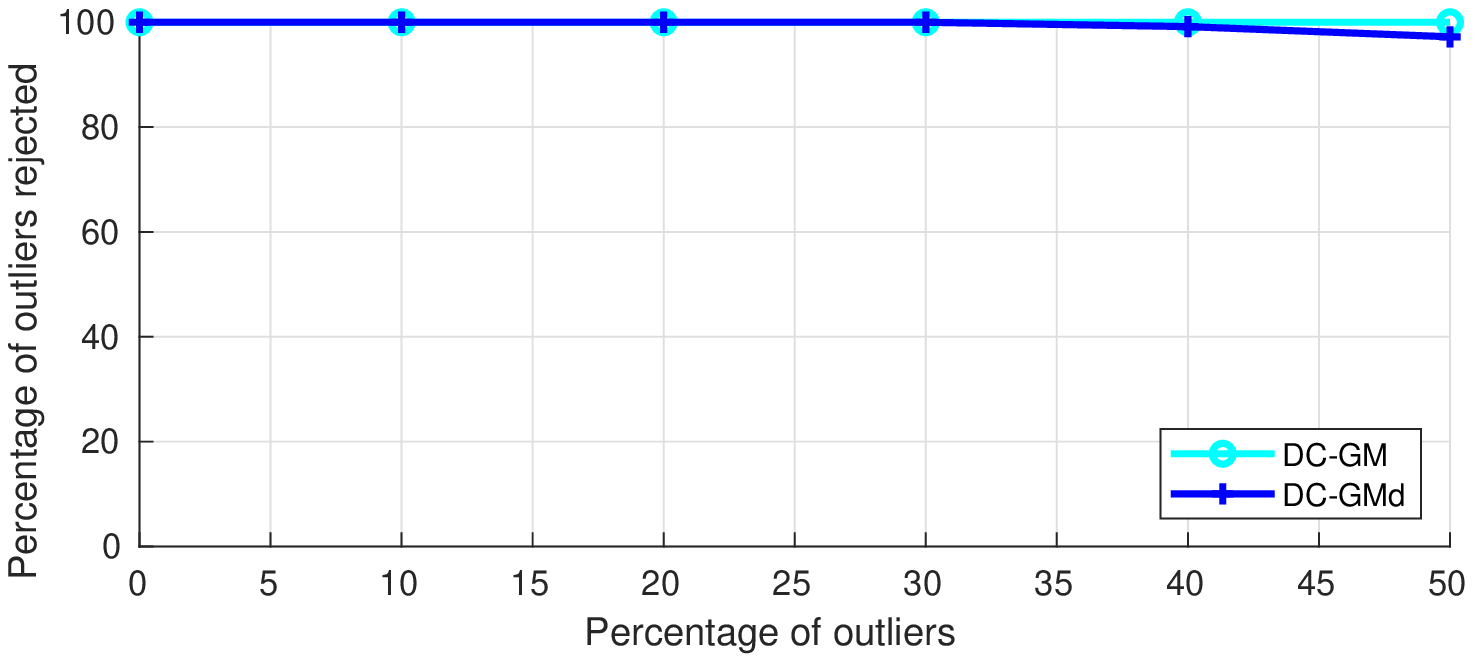} (d)
}
\end{minipage}
\caption{Results on the simulated grid graph with maximum admissible residuals of 1$\sigma$.}
\label{fig:nrsigma1}
\end{figure}
Figure \ref{fig:nrsigma1} presents the results obtained with a threshold of 1 standard deviation on the residuals. Those results have already been discussed in the paper. We observe that for increasing percentage of outliers, \dcMRFc is able to reject all outliers, while \dcMRFd tends to incorrectly accept a very small portion of outliers.
The plot of the rank of the SDP solutions provides some extra insight on the performance of the relaxation and shows that the relaxation of the coupled approach (\dcMRFc) is tighter than the decoupled one (\dcMRFd).

\begin{figure}[H]
\centering
\begin{minipage}{0.4\textwidth}
\centering
\subfloat{
	\includegraphics[width=0.8\textwidth]{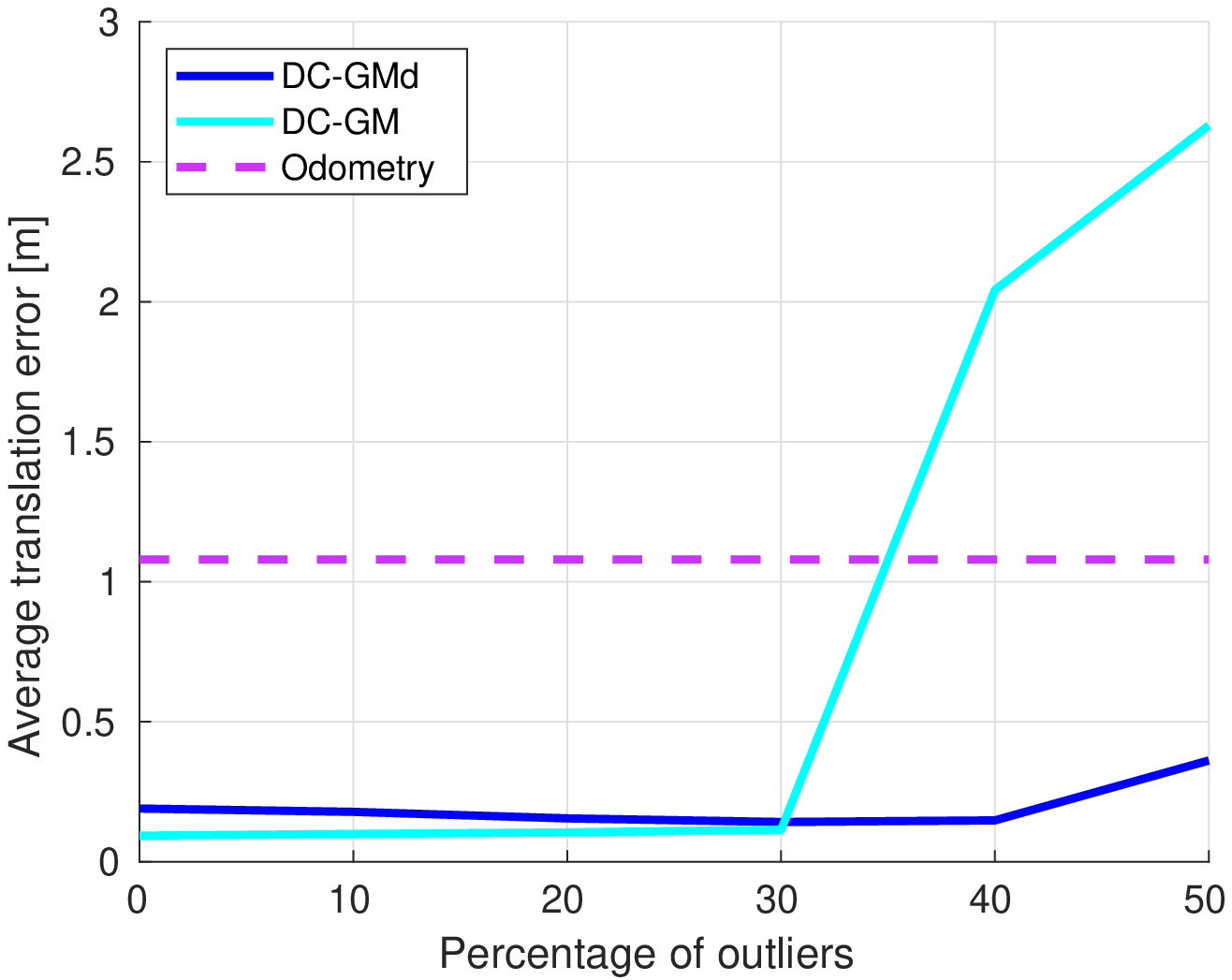} (a)
}

\subfloat{
	\includegraphics[width=0.8\textwidth]{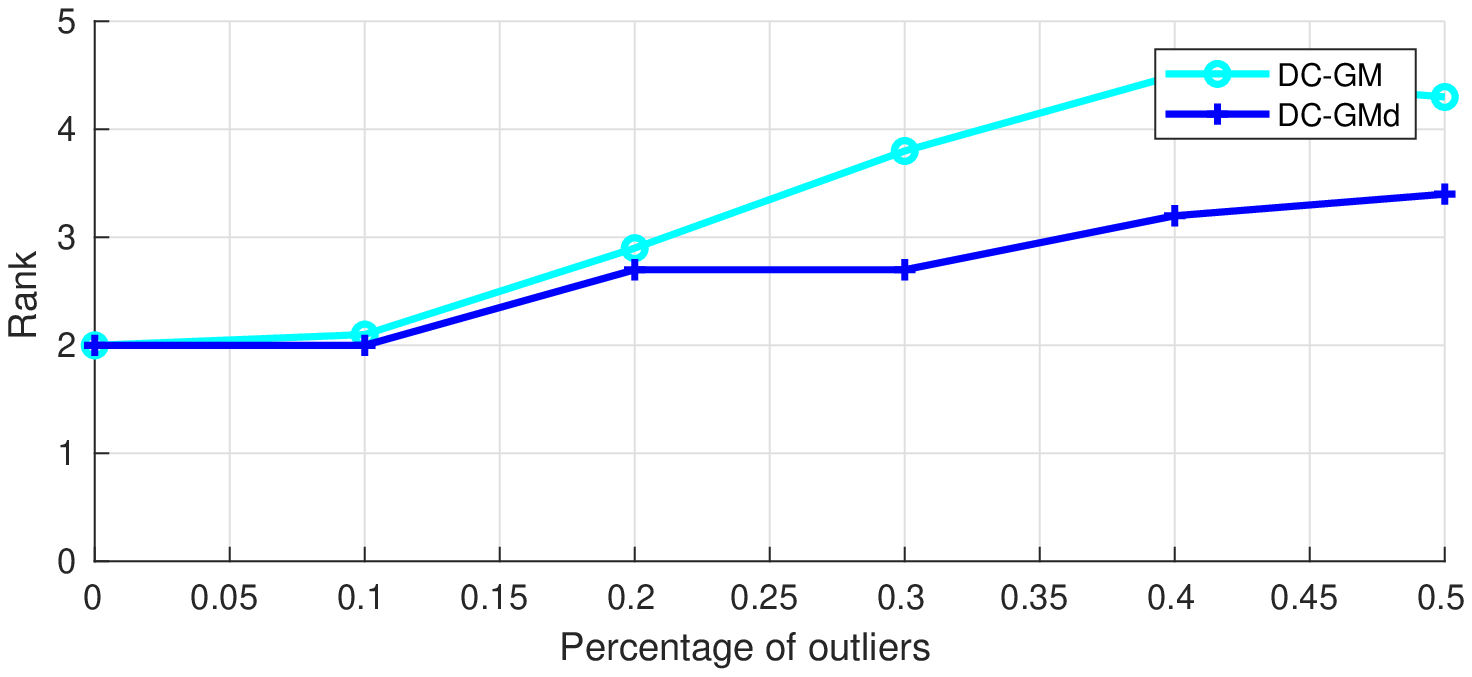} (b)
}
\end{minipage}
\begin{minipage}{0.4\textwidth}
\centering
\subfloat{
	\includegraphics[width=1\textwidth]{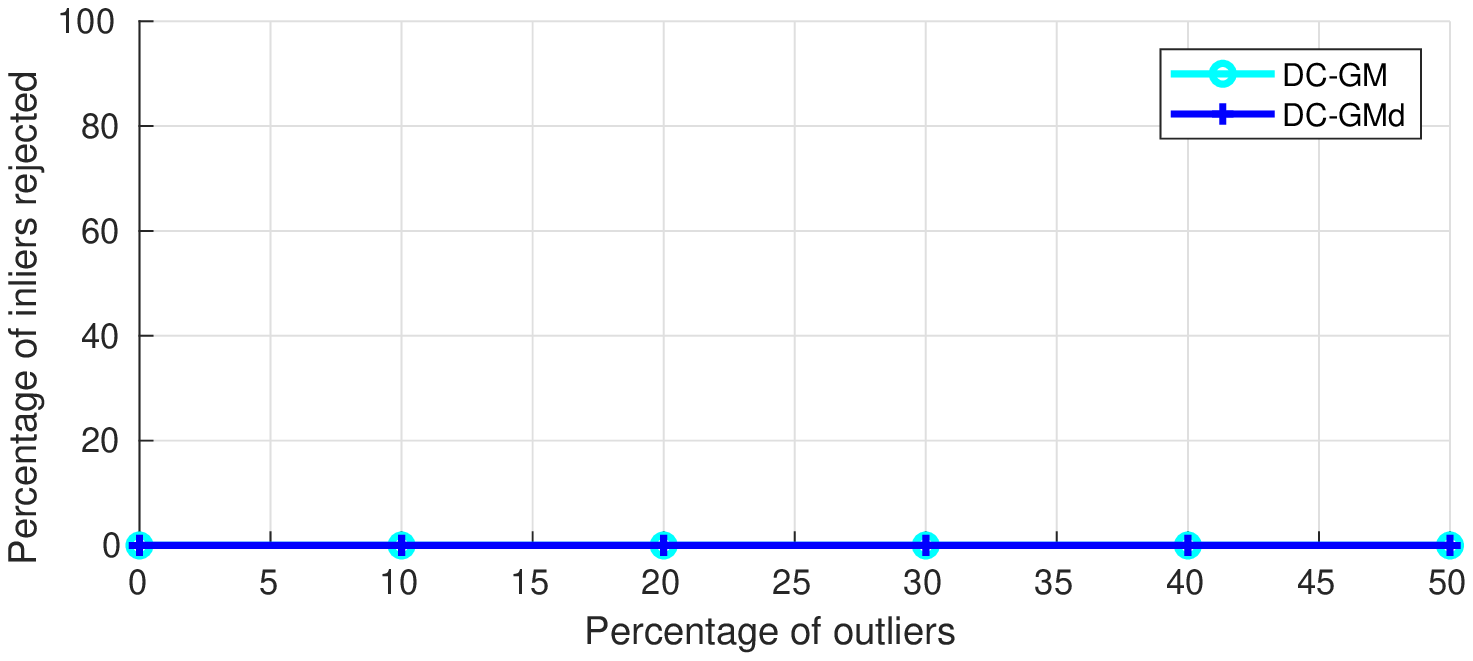} (c)
}
	
\vspace{10mm}
\subfloat{
	\includegraphics[width=1\textwidth]{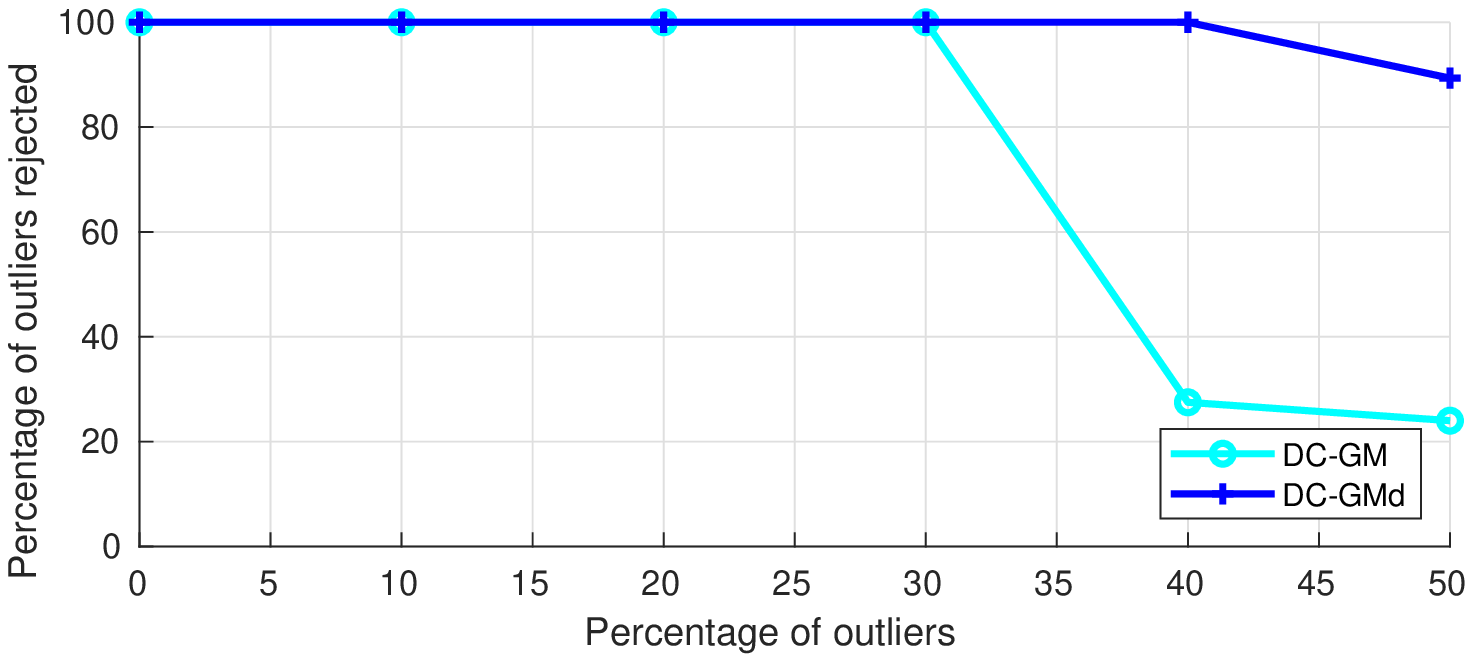} (d)
}
\end{minipage}
\caption{Results on the simulated grid graph with maximum admissible residuals of 2$\sigma$.}
\label{fig:nrsigma2}
\end{figure}
Figure \ref{fig:nrsigma2} shows that with a looser threshold on the maximal residual threshold of 2$\sigma$, the proposed techniques tend to fail in presence of a large amount of outliers (40\% and above).
Surprisingly, the coupled formulation \dcMRFc has a lower breakdown point, and is dominated by \dcMRFd for large percentages of outliers.
Figure \ref{fig:nrsigma2}(c) shows that both techniques accept all the inliers, but Figure \ref{fig:nrsigma2}(d) shows that the loss in accuracy of \dcMRFc stems from accepting several outliers.
This can be partially explained by the 
 the rank in Figure \ref{fig:nrsigma2}(b), which tends to be larger in this case for \dcMRFc, leading to a looser relaxation. 
 A second explanation is provided in Section~\ref{sec:barcij} which shows that the correlation terms have the effect of
 ``inflating'' the maximum admissible residual threshold, making \dcMRFc more prone to accept outliers when $\barc$ is large.
\newpage
\subsection{Effect of the Correlation Terms $\barcij$ on the Objective Function}
\label{sec:barcij}
 

In order to understand the impact of the correlation terms $\barcij$ on the objective function, 
let us consider 
a single loop closure $(i,j)$, 
and call $\calC_{ij}$ the set of edges correlated to $(i,j)$.
 From the coupled formulation, we can isolate all the terms involving the loop closure $(i,j)$ which we report below after omitting constant terms:
 \beal
\;\;\;\;\frac{(1+\bmp_{ij})}{2}\| \resPose \|_\MOmega^2  
& \displaystyle - \frac{\bmp_{ij}}{2} \barc  \;\;- \!\!\!\! \sum_{\;\;\;\;(i',j') \in \calC_{ij}} \barcij \bmp_{ij} \bmp_{i'j'}
\label{eq:robustCostModeling1MeasurementProblem}
\eeal
Now assume that all neighbors ``decide to accept'' the corresponding measurements, i.e., 
$\bmp_{i'j'}=+1$ for all $(i',j') \in \calC_{ij}$. Then, eq.~\eqref{eq:robustCostModeling1MeasurementProblem} becomes:
 \beal
\;\;\;\;\frac{(1+\bmp_{ij})}{2}\| \resPose \|_\MOmega^2  
& \displaystyle 
- \frac{\bmp_{ij}}{2} 
\left( \barc
+ 2\!\!\!\!\!\sum_{\;\;\;\;(i',j') \in \calC_{ij}} \!\!\!\!\!\barcij\right)

\begin{figure}[H] 
\centering	
\includegraphics[width=0.5\textwidth]{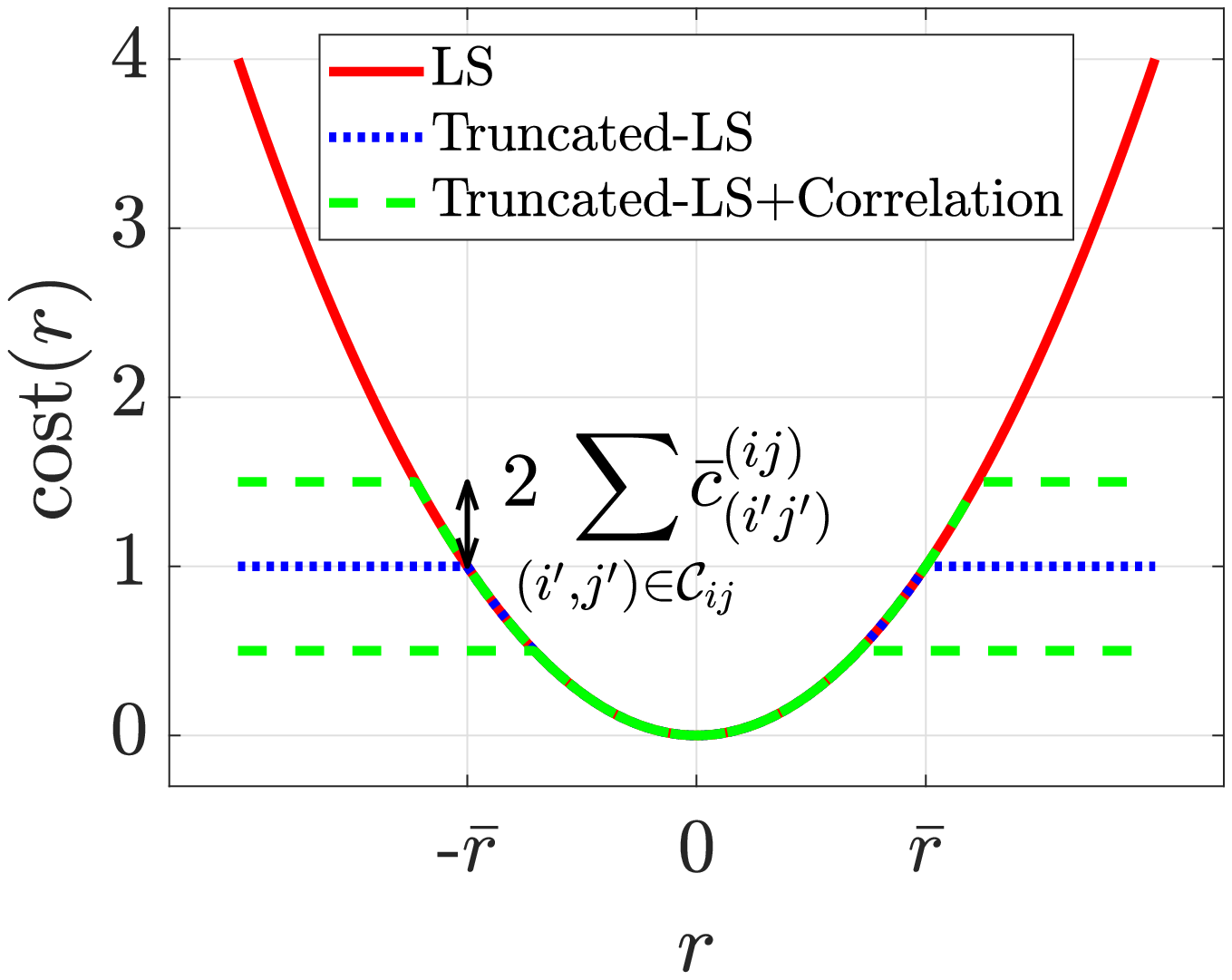}
\caption{Effect of the correlation terms on the robust cost function.}
\label{fig:robustCostModeling}
\end{figure}

\eeal
Similarly, when all neighbors 
``decide to reject'' the corresponding measurements, i.e., 
$\bmp_{i'j'}=-1$ for all $(i',j') \in \calC_{ij}$. Then, eq.~\eqref{eq:robustCostModeling1MeasurementProblem} becomes:
 \beal
\;\;\;\;\frac{(1+\bmp_{ij})}{2}\| \resPose \|_\MOmega^2  
& \displaystyle 
- \frac{\bmp_{ij}}{2} 
\left( \barc - 2\!\!\!\!\!\sum_{\;\;\;\;(i',j') \in \calC_{ij}} \!\!\!\!\!\barcij\right)
\eeal
It is clear that in general, the presence of the correlation term alters the value of the threshold $\barc$. In other words, the fact that neighboring edges accept a measurement, makes the other edges ``more permissive'' by increasing the corresponding threshold $\barc$. The threshold however always remains in the interval:
\beq
\label{eq:interval}
[ \barc - 2\!\!\!\!\!\sum_{\;\;\;\;(i',j') \in \calC_{ij}} \!\!\!\!\!\barcij\quad, \quad \barc + 2\!\!\!\!\!\sum_{\;\;\;\;(i',j') \in \calC_{ij}} \!\!\!\!\!\barcij]
\eeq
A pictorial representation is given in
Figure \ref{fig:robustCostModeling}. This understanding also informs us on how to set the coefficients $\barcij$. According to~\eqref{eq:interval}, one should make sure that $\barcij$ is relatively small compared to $\barc$, such that the correlation does not dominate the outlier rejection decisions. Similarly, the size of the interval in~\eqref{eq:interval} depends on the number of neighbors of edge $(i,j)$; this suggests normalizing the coefficients by the number of neighbors such that the term $\sum_{(i',j') \in \calC_{ij}} \!\barcij$ does not dominate the outlier rejection threshold $\barc$. 

Finally, it is interesting to note that the presence of the correlation terms can degrade the performance of \dcMRFc if the correlation terms are chosen incorrectly. 
We analyze this aspect in the following section where we consider experiments with heterogeneous groups of measurements (i.e., a mix of inliers and outliers). 
\newpage
\subsection{Effect of heterogeneous groups of loop closures}\label{sec:heterogeneous}

The experiments in this section evaluate the impact of an incorrect modeling of the outlier correlation. In particular, 
we  consider a setup where heterogeneous loop closure groups (composed of both inliers and outliers) are added to the graph and 
we add correlation terms $\barcij$ between each pair of edges in the groups. This incorrect modeling is expected to challenge the performance of \dcMRFc, since the model will attempt to encourage consistent inlier/outlier decisions within each group, despite the fact that each edge in the group is assigned to be an inlier/outlier at random.
In particular, we expect \dcMRFc to perform poorly when the correlation term $\barcij$ is large, while it is expected to fall back to the performance of the decoupled approach \dcMRFd when $\barcij$ is small. 
We present results for decreasing value of the correlation term $\bar{c}^{(ij)}_{(i'j')}$ equal to 10\%, 1\%, and 0.1\% of the maximum admissible residuals parameter $\bar{c}$, respectively. The results, showing the percentage of rejected inliers and outliers for increasing percentage of outliers, confirm the expected behavior.

\begin{figure}[H]
\centering
\subfloat{
	\includegraphics[width=0.45\textwidth]{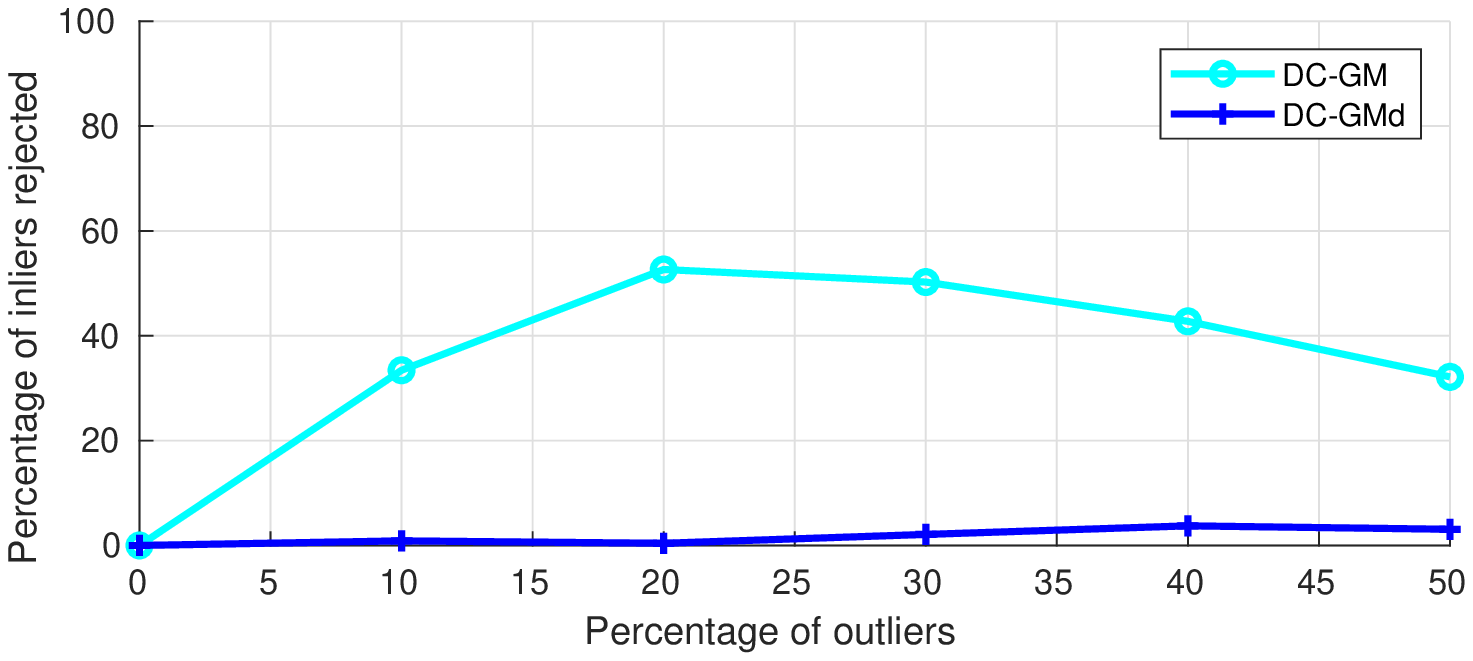} }
\subfloat{
	\includegraphics[width=0.45\textwidth]{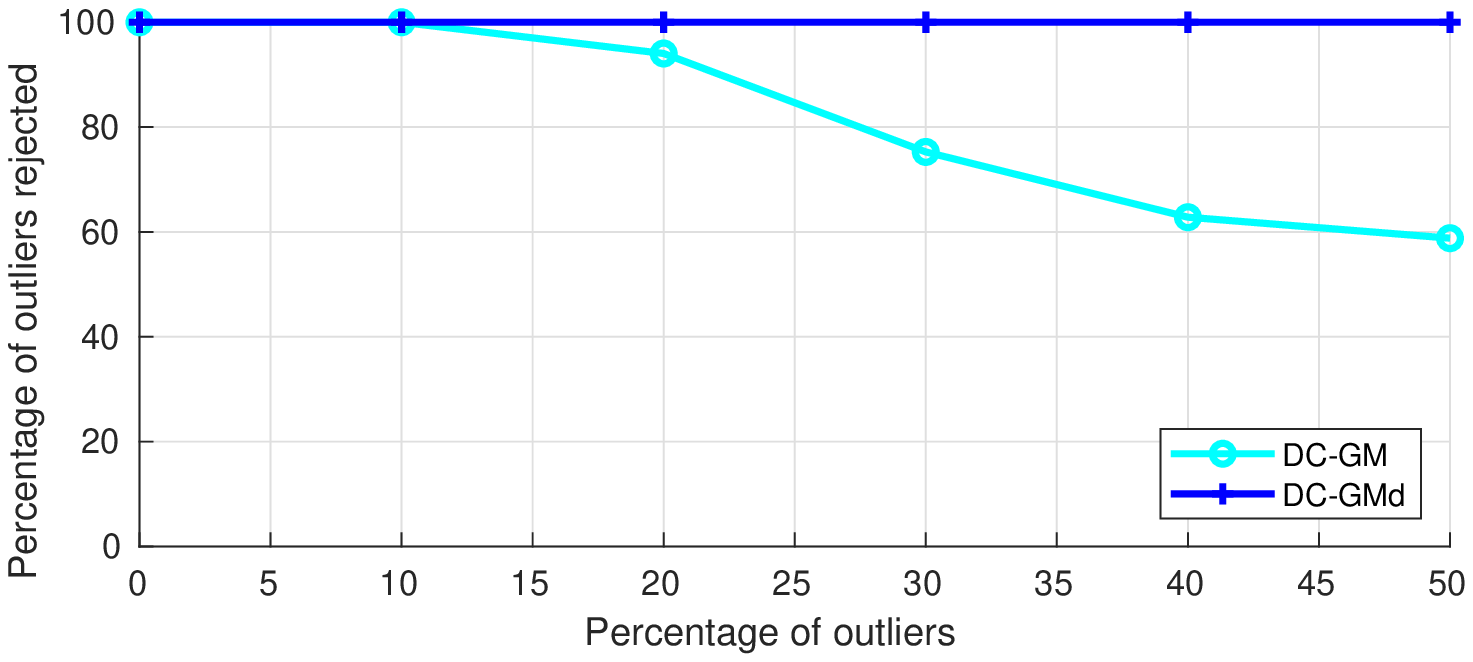} }
\caption{Results on the simulated grid graph with heterogeneous groups of loop closures and correlation terms $\barcij$ equal to $0.1\bar{c}$. (left) Percentage of rejected inliers; (right) Percentage of rejected outliers.}
\label{fig:0_1c}
\end{figure}

\begin{figure}[H]
\centering
\subfloat{
	\includegraphics[width=0.45\textwidth]{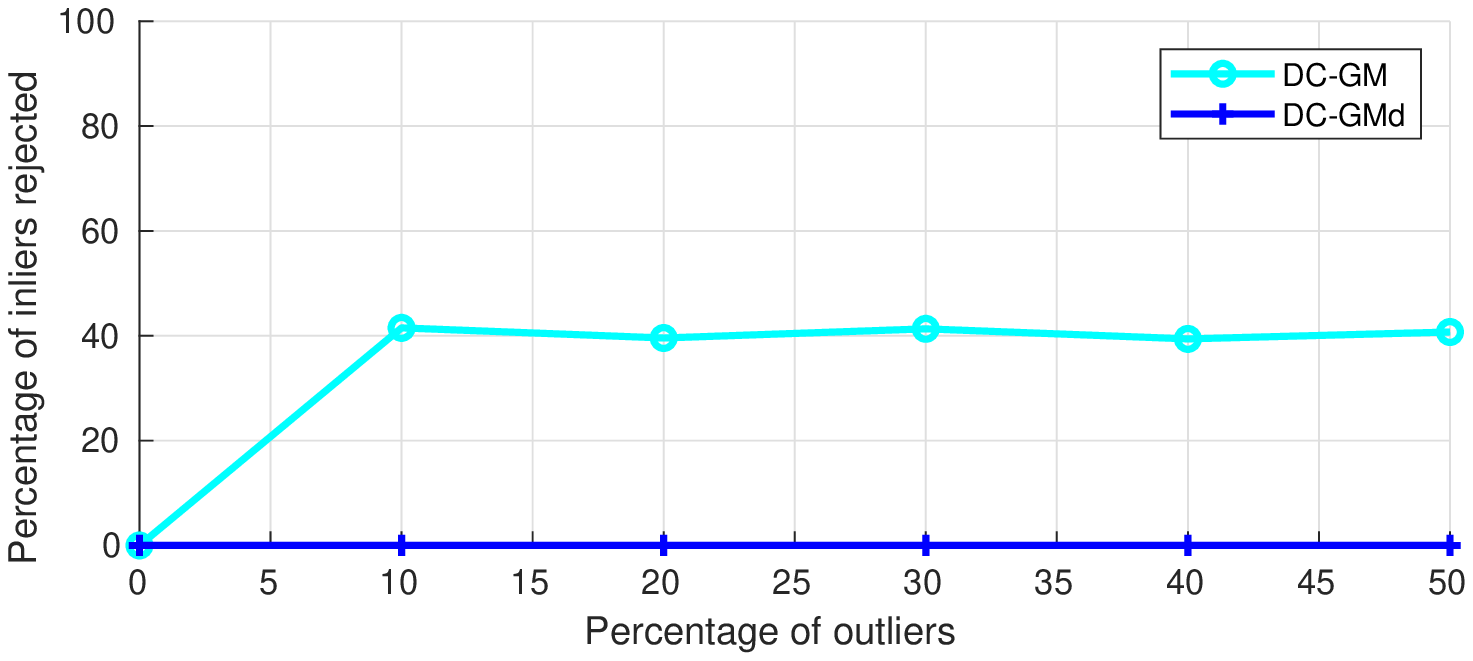} }
\subfloat{
	\includegraphics[width=0.45\textwidth]{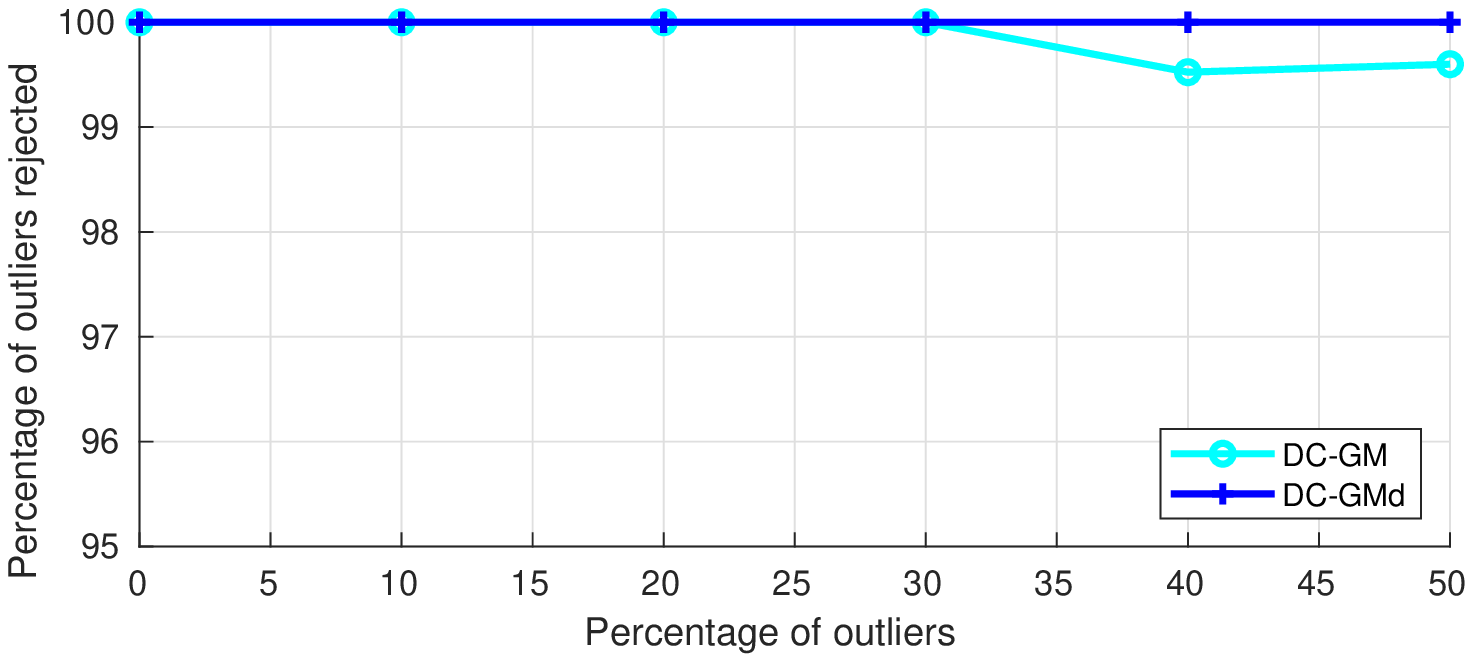} }
\caption{Results on the simulated grid graph with heterogeneous groups of loop closures and correlation terms $\barcij$ equal to $0.01\bar{c}$.}
\label{fig:0_01c}
\end{figure}

\begin{figure}[H]
\centering
\subfloat{
	\includegraphics[width=0.45\textwidth]{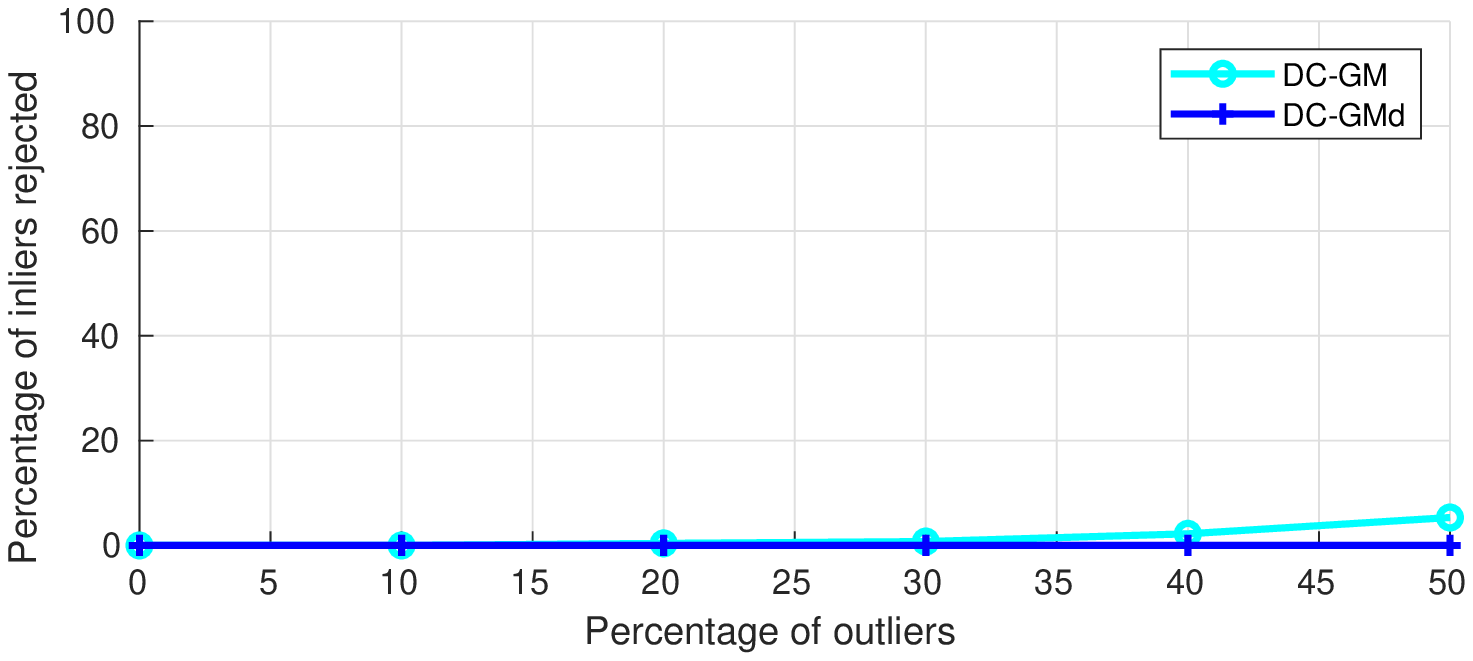} }
\subfloat{
	\includegraphics[width=0.45\textwidth]{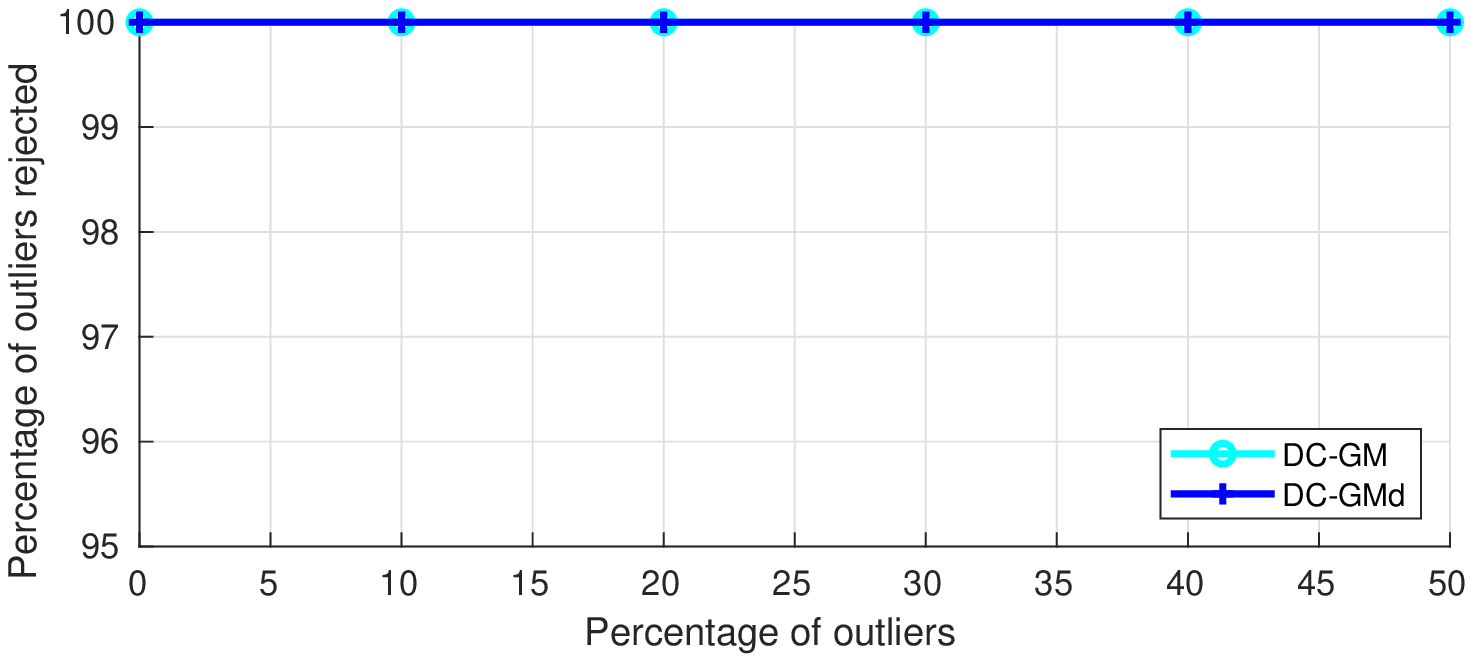} }
\caption{Results on the simulated grid graph with heterogeneous groups of loop closures and correlation terms $\barcij$ equal to $0.001\bar{c}$.}
\label{fig:0_001c}
\end{figure}

Figures \ref{fig:0_1c}, \ref{fig:0_01c}, and \ref{fig:0_001c} show that the performance of \dcMRFc is worse when the correlation terms are large (and incorrect) while it approaches \dcMRFd when the correlation terms are small.
 This is also consistent with the interpretation of the cost function in Figure \ref{fig:robustCostModeling} where  higher values of the correlation terms lead to a larger range of values for the maximum admissible residuals. A larger range is more likely to lead to the acceptance of outliers and/or the rejection of inliers.
\newpage
\subsection{Additional simulation results}\label{sec:partial_grid}
We performed additional simulation experiments to provide further insights to the reader on the performance of the proposed techniques. These experiments involve a more realistic Manhattan World graph. 
Below we report the average translation error, the percentage of rejected inliers and outliers, and a visualization of the estimated trajectory. Statistics are computed over 10 runs with increasing percentage of outliers. The proposed approaches (\dcMRFc and \dcMRFd) are compared against other techniques (Vertigo, DCS, RRR) potentially reporting multiple choices of parameters for the competing techniques.

\begin{figure}[!htbp]
\centering
\begin{minipage}{0.45\textwidth}
\centering
\subfloat{
	\includegraphics[width=1\textwidth]{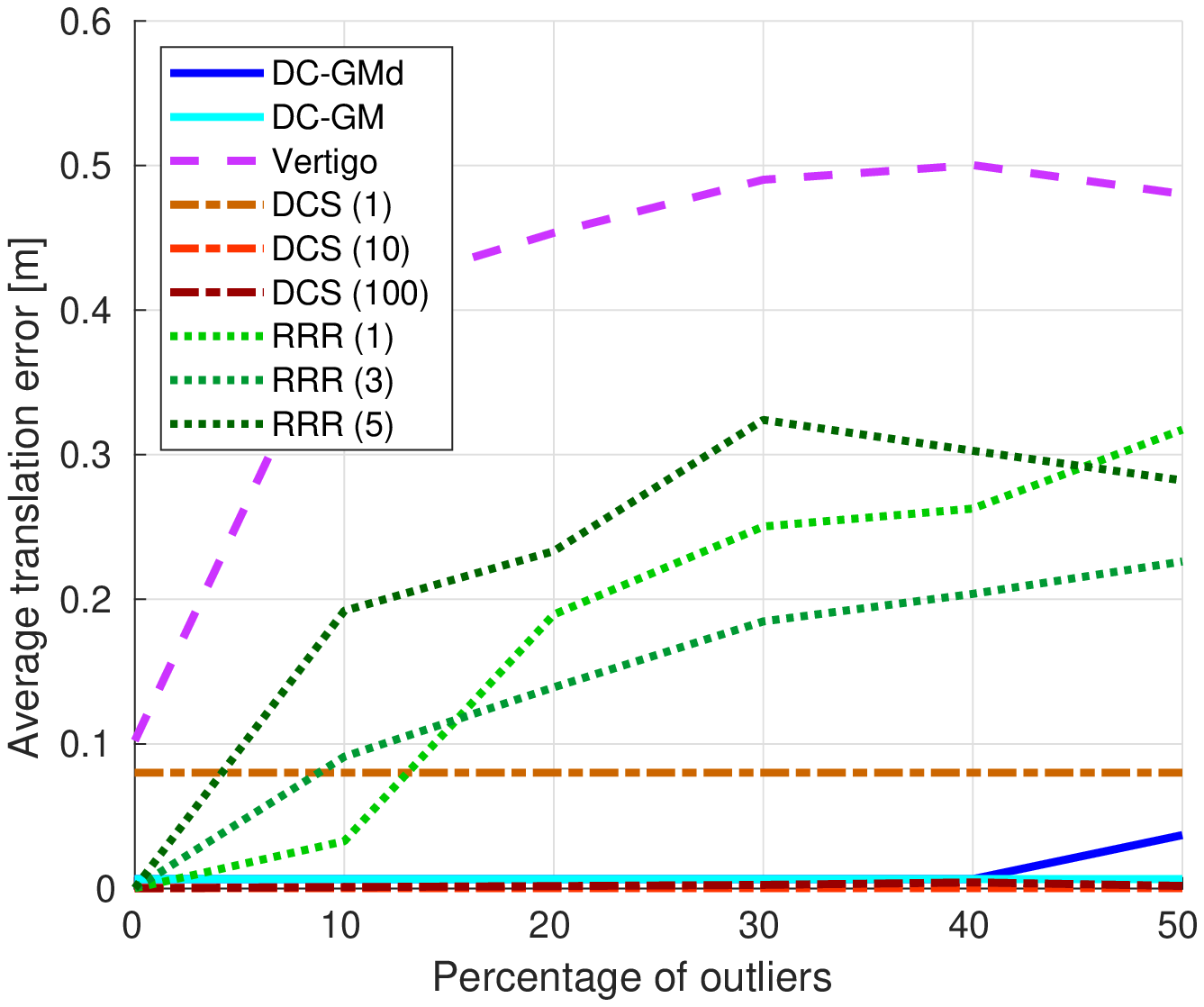} 
	(a)
}

\subfloat{
	\includegraphics[width=1\textwidth]{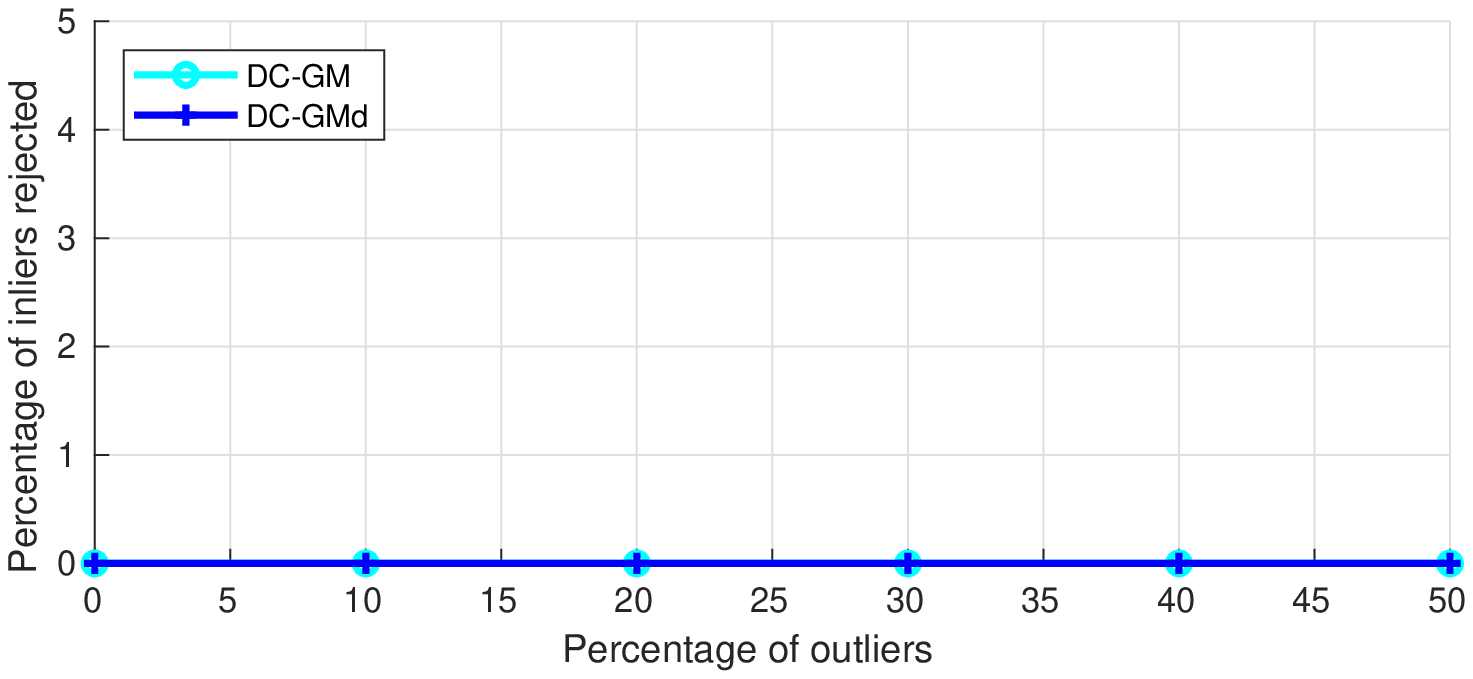}
	(b)
}
\end{minipage}
\hspace{1cm}
\begin{minipage}{0.45\textwidth}
\centering
\subfloat{
	\includegraphics[width=1\textwidth]{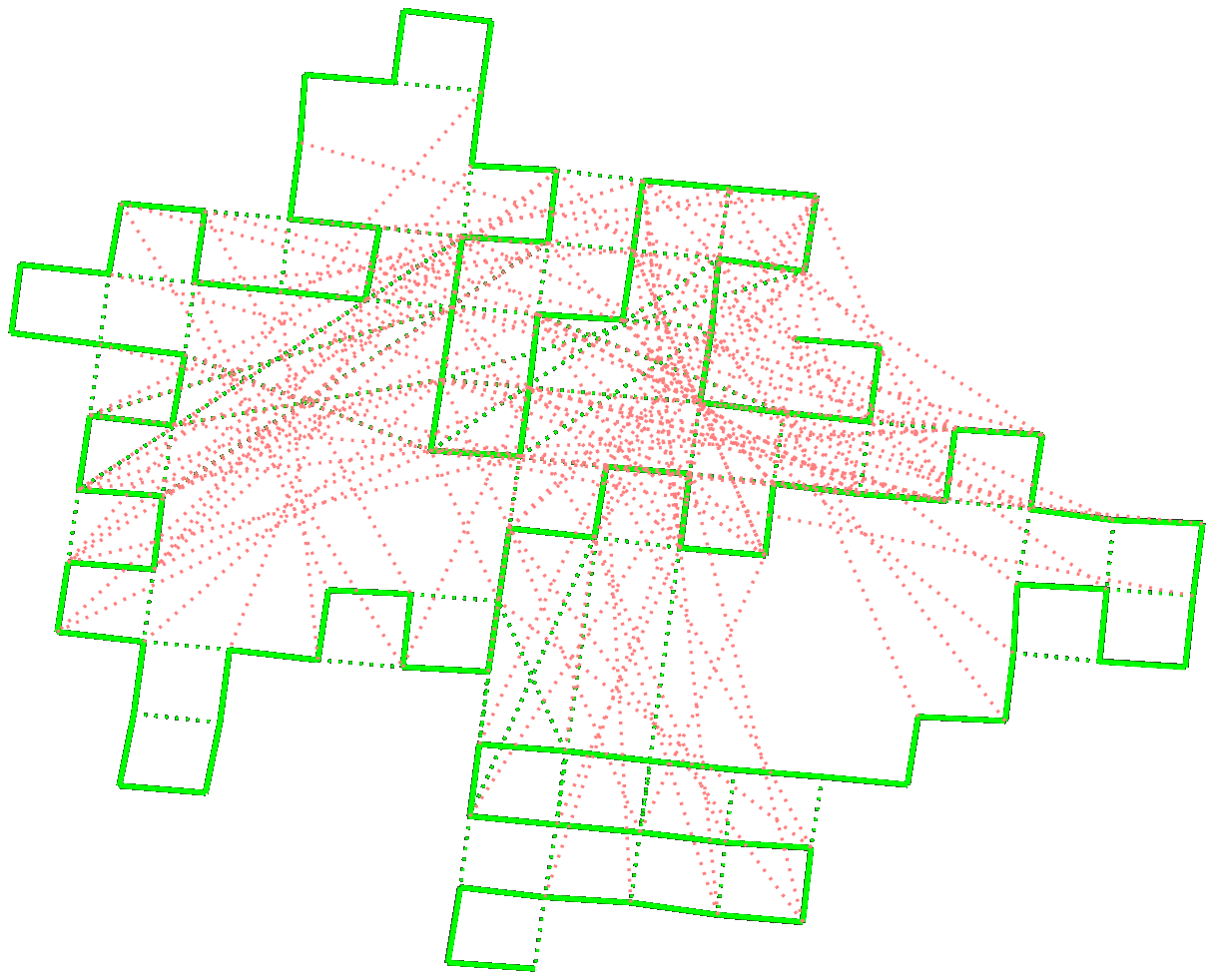} 
	(c)
}

\subfloat{
	\includegraphics[width=1\textwidth]{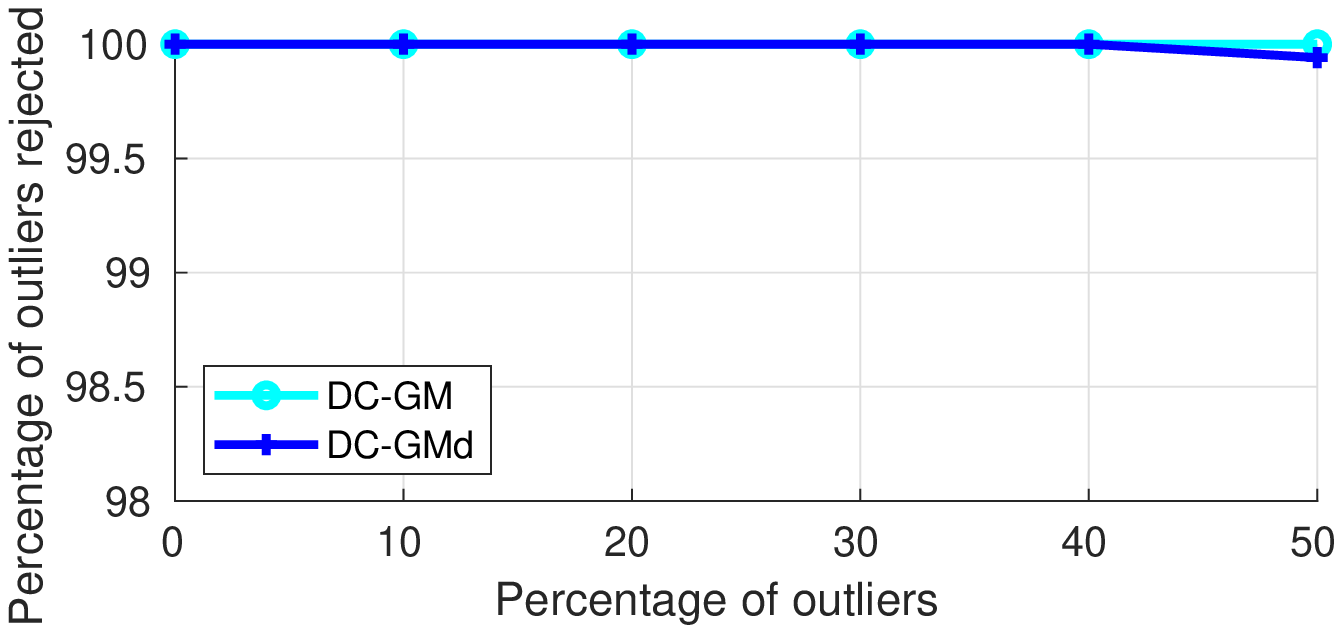} 
	(d)
}
\end{minipage}
\caption{Results on a Manhattan World graph. 
(a) Average translation error for the different techniques;
(c) Ground truth (green) overlaid on the \dcMRFc solution (black, indistinguishable from the ground truth), and outlier loop closures (red).
(b) Percentage of rejected inliers;
(d) Percentage of rejected outliers. }
\label{fig:partial_grid_graph}
\end{figure}

Consistently with the results on the grid graph, Figure~\ref{fig:partial_grid_graph} shows that the performance of \dcMRFc and \dcMRFd is comparable, but \dcMRFc ensures slightly more accurate results when the percentage of outliers is large. 
 On the other hand, both Vertigo and RRR performed remarkably worse than 
\dcMRFc and \dcMRFd on the Manhattan World graph (for RRR the performance was poor for any choice of parameters).
DCS performed well when the tuning parameter $\Phi$ was chosen to be $10$ or $100$, but performed worse than \dcMRFc when the default parameter $\Phi=1$ was used.




\end{document}